\documentclass[letterpaper, 10 pt, conference]{ieeeconf}  % Comment this line out if you need a4paper

\IEEEoverridecommandlockouts                              % This command is only needed if 
\usepackage{cite}

\usepackage{amsmath,amssymb,amsfonts,amsthm}
\usepackage{algorithmic}

\usepackage{graphicx}
\usepackage{textcomp}
\usepackage{hyperref}
\usepackage{caption}
\usepackage{soul}
\usepackage{subcaption}
\usepackage[dvipsnames]{xcolor}
\usepackage{glossaries}
\newacronym{DRL}{DRL}{Deep Reinforcement Learning}
\newacronym{MDP}{MDP}{Markov Decision Process}
\newacronym{DDPG}{DDPG}{Deep Deterministic Policy Gradient}
\newacronym{PID}{PID}{Proportional-Integral-Derivative}
\newacronym{PMC}{PMC}{Probabilistic Model Checking}
\newacronym{DTMC}{DTMC}{Discrete Time Markov Chain}
\newacronym{CTMC}{CTMC}{Continuous Time Markov Chain}
\newacronym{LTL}{LTL}{Linear Temporal Logic}
\newacronym{PCTL}{PCTL}{Probabilistic Computational Tree Logic}
\newacronym{RAS}{RAS}{Robotics and Autonomous Systems}
\newacronym{DNN}{DNN}{Deep Neural Networks}
\newacronym{vnv}{V\&V}{Verification and Validation}

\newcommand{\next}{\bigcirc}
\newcommand{\eventually}{\Diamond}

\def\BibTeX{{\rm B\kern-.05em{\sc i\kern-.025em b}\kern-.08em
    T\kern-.1667em\lower.7ex\hbox{E}\kern-.125emX}}

\usepackage{comment}

\newtheorem{remark}{Remark}
\newtheorem{definition}{Definition}
\usepackage[absolute,overlay]{textpos}% by XZ: for the text in the top margin..

\usepackage{color}
\newcount\Comments  % 0 suppresses notes to selves in text
\definecolor{darkgreen}{rgb}{0,0.5,0}
\definecolor{purple}{rgb}{1,0,1}
\newcommand{\kibitz}[2]{\ifnum\Comments=1\textcolor{#1}{#2}\fi}
\title{\LARGE \bf Dependability Analysis of Deep Reinforcement Learning based Robotics and Autonomous Systems through Probabilistic Model Checking %\vspace{-4mm}% XZ: A candidate...
}
\author{Yi Dong$^{1}$, Xingyu Zhao$^{1}$ and Xiaowei Huang$^{1}$% <-this % stops a space
\thanks{$^{1}$Department of Computer Science, University of Liverpool, UK
        {\tt\small \{yi.dong,xingyu.zhao,xiaowei\}@liverpool.ac.uk}}%
}

\begin{document}

%%%%%%%%added by XZ%%%%%%%%%%
\begin{textblock*}{20cm}(3cm,1cm)
	\textcolor{red}{        Preprint accepted by IROS2022. To appear in IROS2022 proceedings on IEEE Explore}.
\end{textblock*}
%%%%%%%%end by XZ%%%%%%%%%%

\maketitle
\thispagestyle{plain}
\pagestyle{plain}
\begin{abstract}
While Deep Reinforcement Learning (DRL) provides transformational capabilities to the control of Robotics and Autonomous Systems (RAS), the black-box nature of DRL and uncertain deployment environments of RAS pose new challenges on its dependability. Although existing works impose constraints on the DRL policy to ensure successful completion of the mission, it is far from adequate to assess the DRL-driven RAS in a holistic way considering all dependability properties. In this paper, we formally define a set of dependability properties in temporal logic and construct a Discrete-Time Markov Chain (DTMC) to model the dynamics of risk/failures of a DRL-driven RAS interacting with the stochastic environment. We then conduct Probabilistic Model Checking (PMC) on the designed DTMC to verify those properties. Our experimental results show that the proposed method is effective as a holistic assessment framework while uncovering conflicts between the properties that may need trade-offs in training. Moreover, we find that the standard DRL training cannot improve dependability properties, thus requiring bespoke optimisation objectives. Finally, our method offers sensitivity analysis of dependability properties to disturbance levels from environments, providing insights for the assurance of real RAS.
\end{abstract}

% \begin{IEEEkeywords}
% Dependability Analysis, Deep Reinforcement Learning, Robotics and Autonomous Systems, Probabilistic Model Checking
% \end{IEEEkeywords}
% % \vspace{-1mm}
\section{Introduction}

The major obstacle to reaping the benefits of \gls{RAS} is the assurance of their dependability \cite{lane_new_2016}---an umbrella concept {holistically} covering aspects of a system’s quality, including reliability, safety, availability, and performance \cite{avizienis_basic_2004}.  Since first proposed in 2013 \cite{mnih2013playing}, \gls{DRL} has received significant attention in many applications \cite{dong2021strategic,sallab2017deep,cheng2019end}. DRL, yielding a control policy for \gls{RAS}, is replacing the traditional control algorithms, thanks to its ability to deal with complex and nonlinear problems \cite{hu2020voronoi,schoettler2020deep,li2018off}.
However, the existing research on safe \gls{DRL} mainly focuses on eliminating unsafe ``state-action'' pairs that may cause failures \cite{behzadan2019adversarial,alshiekh2018safe,jansen2018shielded,zhang2020robust}, without considering more involved dependability properties, including robustness, resilience and safe requirements \cite{bloomfield2020towards}. For example, a policy that consistently leads the vehicle to go in circles is safe but unacceptable; meanwhile, a policy that recovers quickly from an unsafe state is better than a slower strategy.  
There is an urgent need to (1) consider a set of  dependability properties that concerns not only the completion of missions but also the quality of the completions, and (2) develop methods to evaluate and certify the dependable use of DRL-driven RAS in critical applications \cite{robu_train_2018}. This paper addresses this need through \gls{PMC}. First, a risk-aware \gls{DTMC} is constructed to model the interactions of the DRL agent with stochastic environments during the execution. Second, an off-the-shelf \gls{PMC} tool is applied to the DTMC to verify a set of \emph{quantitative dependability properties} expressed in the temporal logic \gls{PCTL}.

\begin{figure}[htbp]
\centerline{\includegraphics[width=\hsize]{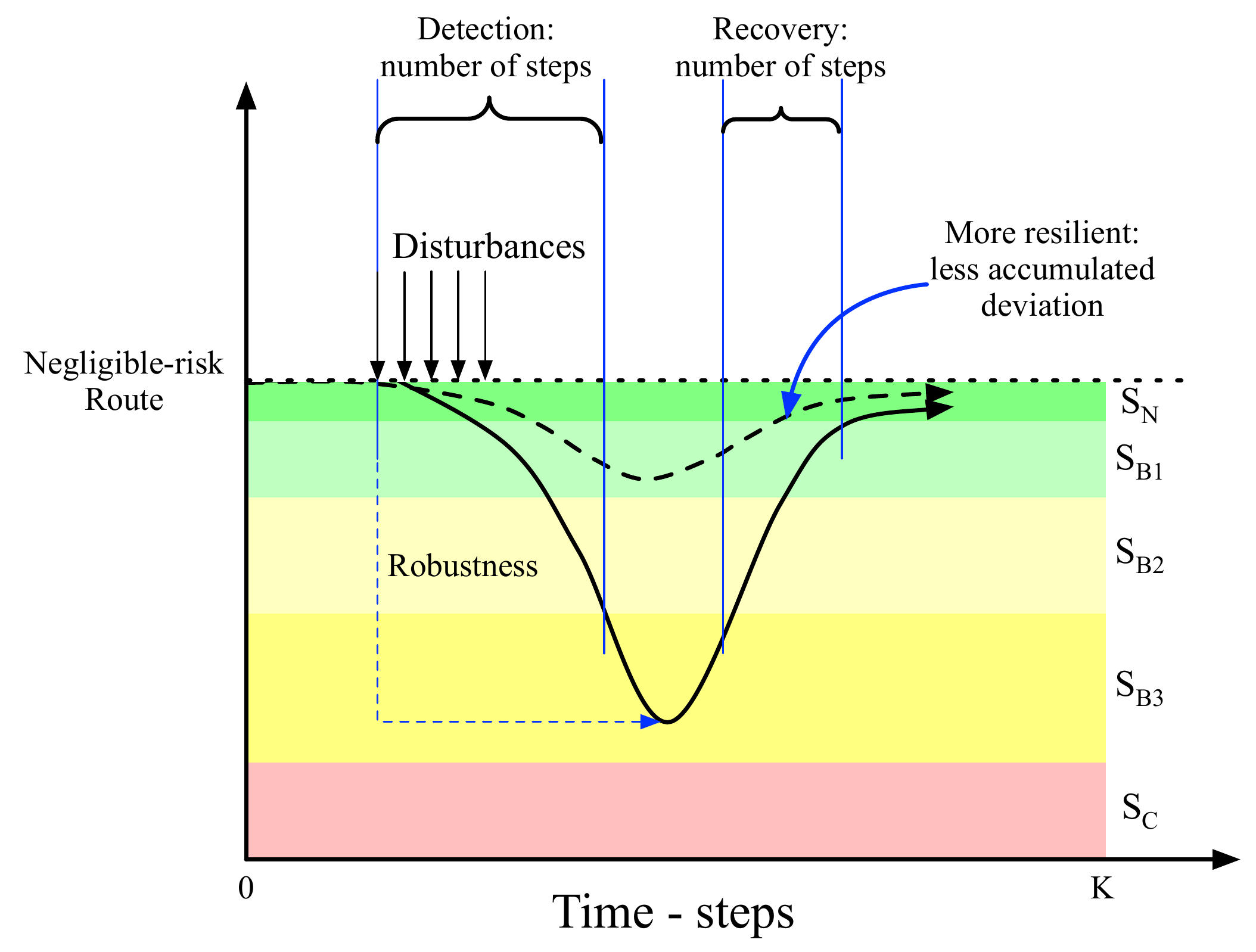}}
%\vspace{-0.5em}
\caption{Conceptualised illustration of dependability properties: safety, resilience, robustness, detection and recovery. Colour (red to green) indicates risk levels (high to low). 
}
\label{properties}
\end{figure}

As illustrated in Fig.~\ref{properties}, we consider, in addition to the safety (i.e., completion of mission without failures), other dependability properties, including robustness, resilience, detection, and recovery.  Simply speaking, in an environment where the robot's sensory input may be subject to disturbances, 
\emph{robustness} expresses the ability to complete the mission regardless of the disturbance,
%recover from risky situations\xingyu{Is this consistent with what we defined later? The mission completion (classification) is not affected by disturbance (perturbation).. },
\emph{resilience} evaluates the accumulated deviation from a negligible-risk trajectory, \emph{detection} concerns how fast the robot's risky situation may deteriorate (and therefore be detected), and \emph{recovery} concerns how soon the robot can recover from a risky situation. 
%The aforementioned properties are defined quantitatively in \gls{PCTL}, and our framework can work with any property expressible with PCTL\xingyu{to chat...}. 
Different from the safe RL research, \cite{JMLR:v16:garcia15a} which mainly concerns the reachability of error states or bad events, the above properties require explicit consideration of a sequence of states and the states on another path (i.e., the negligible-risk trajectory as in Fig.~\ref{properties}).
% \xingyu{Indeed, this is our unique feature, we shall have another sentence explaining why this feature is better/desirable than [16]...}
Arguably, such a \textit{holistic} evaluation of a set of quantitative properties is needed to have an in-depth understanding of the robot's (and the DRL's) dependability.

Given the black-box nature of DRL and that we usually do not have a formal model for the environment, it is unlikely that we can have a probabilistic model that captures all the executions. To facilitate a formal analysis, we construct a DTMC from a set of sampled trajectories that can be augmented with domain knowledge \cite{calinescu_efficient_2021} and other \gls{vnv} evidence \cite{zhao_probabilistic_2019}. The DTMC is dedicated to risk analysis to include only states that represent different levels of risks.

In addition to evaluating the dependability of RAS, we apply our method to study the real \gls{RAS}.
% In addition to evaluate the dependability of RAS in general, we apply our method to study the sim-to-real challenge \cite{9308468}, one of the major challenges of DRL that concerns the difficulty of transferring an DRL agent trained in simulation environment to real environment. 
In particular, we do sensitivity analysis on dependability properties with respect to the disturbance levels of the environment and utilise the results to provide insights on whether, in a given natural environment, a certain level of dependability of the RAS can be achieved after deployment.

In summary, the key contributions of this paper include:
\begin{enumerate}
    \item A set of formally defined dependability properties that need to be evaluated before deploying the DRL-driven RAS in critical applications. %XZ: not just safety critical, also business critical, etc..
    \item An initial framework on constructing failure process DTMCs that model the dynamics of risky situations in executing a DRL-driven RAS.
    %To assess the different properties of the DRL strategy for RAS, a novel multi-level and multi-perspective DTMC of the RAS failure process is proposed.
    
    \item A publicly accessible repository of our proposed method with all source code, datasets and experimental results (including a real-world case study based on Turtlebot Waffle Pi). 
    
    %An implementation of the proposed method are open-sourced, and the method is deployed in a real-world autonomous system based on Turtlebot Waffle Pi, together with a practical study on the DRL. \xingyu{a link to your github repo?}
    % \item \textcolor{blue}{The proposed method considers the dependability of RAS based on trajectories, which are better than single state analysis in the concept of DRL.}
\end{enumerate}

% \textcolor{blue}{The rest of this paper is organised as follows. The related works of this paper are summarised in Section \ref{sec.2}. 
% Preliminaries are presented in Section \ref{sec.3}.
% The formulation of autonomous vehicle and robots are presented in Section \ref{sec.4}, including running examples and \gls{DTMC}. Then, several formal properties are designed for robotics and autonomous systems in Section \ref{sec.5}. Simulation results and corresponding analysis are presented in Section \ref{sec.6}. Finally, Section \ref{sec.7} concludes this paper.}

% \vspace{-1mm}
\section{Related Works}\label{sec.2}

%The earlier methods to 
Most research in safe DRL focuses on enhancing safety and robustness by reducing  potential unsafe actions, including methods for safe monitoring and adversarial training. 
Safe action sets are designed to avoid the unsafe states only based on the current state of the agent, including Shield \cite{alshiekh2018safe,jansen2018shielded}, Lyapunov method \cite{berkenkamp2017safe,huh2020safe}, etc. 
For example, shielding methods prevent agents from making unsafe actions at each state. 
Although choosing an action from the bounded safe action set can return a safe action for that specific time, the correctness of the action at any specific time depends on the expected long-term accumulated rewards. For this reason, the verification of a DRL agent also needs to consider the current state and long-term rewards (and therefore, the future states). 
Mandlekar \textit{et al.} \cite{mandlekar2017adversarially} used actively chosen adversarial perturbations for robust policy training to improve robustness (resistance to changes) in complex environments.
%where analytic calculations are impractical. 
In \cite{bansal2017emergent} and \cite{kurach2020google}, it is found that the DRL agent and adversarial agent can be trained in a cyclical way to avoid overfitting. 
Here, the mentioned manners do not consider the environment model. 
Therefore, to understand if a learned policy works well in an environment, we can conduct a \gls{PMC} on their induced DTMC. This enables the analysis of various properties that can be expressed with PCTL. 
% Here, the mentioned training manner is based on a single agent reinforcement learning algorithm since it does not train both agents in a same time.

In addition, these methods cannot ascertain whether a DRL model satisfies specific properties with provable guarantees. Verification techniques are required for this purpose, but unfortunately, due to the complexity of verification problems (NP-complete for robustness verification \cite{katz2017reluplex,RHK2018} over \gls{DNN}), a direct verification is suffering from the scalability issue and can only work with miniature models. Compared to DNN verification, \gls{DRL} verification is more complicated because it requires the consideration of not only the learned model but also the environment,  which in general does not have a formal model.  
%Therefore, 
% {To make the \gls{DRL} verification practical}, an intuitive idea is to use an 
%equivalent 
% approximate model to replace the policy network %equivalently 
An intuitive idea to make the DRL verification practical is to use an approximate model to replace the policy network \cite{fulton2018verifiably}. 
%This method has high requirements for the equivalent model, and at least the following two conditions must be met: (1) The performance of the equivalent model can be at the same level as the original strategy (or slightly weaker); (2) The equivalent model is easier to be verified its properties, such as safety, robustness and resilience. Therefore, 
For example, a decision tree based approximation model has been considered in \cite{bastani2018verifiable}. 
Behzadan \textit{et al.} proposed a new framework based on \gls{DRL} to benchmark the behaviour of the collision avoidance mechanism in the worst case \cite{behzadan2019adversarial}. They verified the effectiveness of the framework by comparing the reliability of two collision avoidance mechanisms in dealing with deliberate collision attempts over, e.g.,  the number of collisions, return values, and the time from the start to the collision. 
%For example, the longer time indicates that this policy has a stronger anti-collision ability. 
% \st{Kazak \textit{et al.} applied bounded model checking method on the verification of DRL, by unfolding a DRL into a finite number of steps and then conducting DNN verification on individual steps} \cite{10.1145/3341216.3342218}. 
Although these works are building equivalent models to replace the DRL models, not all dependability properties are covered like our method.

Based on traditional software systems, Zhu \textit{et al.} developed the formal verification technology for reinforcement learning verification \cite{zhu2019inductive}. The proposed verification toolchain can ensure that the RL-based control policies are safe in terms of an infinite state transition system specification.%The proposed verification toolchain can ensure that the RL based control policies are safe in terms of a specification of an infinite state transition system.
%that captures technology does not use the black box method to check and change the structure of the neural network to enhance security. Fitting strategy, and then get a simpler and more explanatory synthetic program. Through counterexamples and syntactically guided induction synthesis process to solve the neural network verification problem, and use a verification process to ensure that the state proposed by the program is always consistent with the original specification It is consistent with the inductive invariant of the deployment environment context. This invariant defines an inductive attribute that separates all reachable (safe) and unreachable (unsafe) states that can be expressed in the conversion system.
In addition, to evaluate the robustness and resilience of the agent in the test phase against adversarial disturbances in a way independent of the attack type, Behzadan \textit{et al.} proposed to measure the resilience and robustness of \gls{DRL} strategies. 
%First define the concept of adversarial regret \cite{behzadan2019rl}. 
Different from the above, we define a set of quantitative dependability properties and apply a \gls{PMC} on a failure process DTMC.

%Adversarial regret refers to the difference between the return obtained by the undisturbed subject at time T and the return obtained by the interfered subject at time T, then elasticity refers to the need to cause the greatest adversarial regret The minimum number of disturbance states, robustness refers to the maximum adversarial regret that can be achieved given the maximum number of disturbances. Through the experimental evaluation on DQN, A2C and PPO2 agents trained in the Cart-Pole environment, DQN is in With a small number of disturbance states, the same amount of adversarial regret is caused, indicating that its elasticity is poor, followed by the PPO2 strategy, and the elasticity of the A2C strategy is the strongest of the three. For the maximum of 10 disturbance states In this case, the robustness of the three is very close. This is because the most appropriate number of disturbance states is 7.5 in the calculation of elasticity. When the maximum of is 5 perturbation states, DQN has the largest adversarial regret value, indicating that its robustness is the worst, while A2C has a smaller adversarial regret value, indicating the strongest robustness.
% \vspace{-1mm}
\section{Preliminaries}\label{sec.3}

\subsection{Interaction of Robot with Environment}\label{sec:mdp}

%\gls{DRL} algorithms have been widely used in robotics applications thanks to its ability to learn intelligent agents in complex environments. \gls{DRL} algorithms are concerned with how the robot should act in an environment, and decide a strategy to maximise the trajectory rewards. Therefore, the

We use discounted infinite-horizon \gls{MDP} to model the interaction of an agent with the environment $E$. An MDP is a 5-tuple ${\cal M}^E=(\mathcal{S},\mathcal{A}, \mathcal{P}, \mathcal{R}, \gamma)$, where $\mathcal{S}$ is the state space, $\mathcal{A}$ is the action space, $\mathcal{P}(s'|s,a)$ is a probabilistic  transition, $\mathcal{R}(s,a)\in {\mathbb R}_{\ge 0}$ is a reward function, and $\gamma\in [0,1)$ is a discount factor. A (deterministic\footnote{We consider \gls{DDPG} \cite{lillicrap2015continuous,sutton2018reinforcement,mnih2013playing} for a reinforcement learning agent. DDPG returns a deterministic policy.}) policy is $\pi: \mathcal{S} \rightarrow \mathcal{A}$ that maps from states to actions. 

Based on $\mathcal{M}^E$, a policy $\pi$ induces a trajectory distribution $\rho^{\pi,E}(\zeta)$ where $\zeta=(s_0,a_0,s_1,a_1,...)$ denotes a random trajectory. The state-action value function of $\pi$ is defined as $Q^{\pi}(s,a)= \mathbb{E}_{\zeta\sim \rho^{\pi,E}}[\sum_{t=0}^{\infty}\gamma^t\mathcal{R}(s_t,a_t)]$ and the state value function of $\pi$ is $V^\pi(s)=Q^{\pi}(s,\pi(s))$. 
In Section~\ref{sec.3}, we will explain how to construct a DTMC to approximate $\rho^{\pi,E}(\zeta)$. 

There is always some noise in the environment, which will affect the robot's perception of the environment. For each sensor signal $o_t^i \in s_t$, the existence of the disturbances suggests that the actual sensor reading may be deviated from its value in the current state. Formally, we assume that the actual state $\hat{s}_t$ is within certain norm distance from $s_t$, i.e., $||\hat{s}_t - s_t||_p \leq d$ for some $d>0$, where $||\cdot||_p$ denotes the $p$-norm.%We write $\pi^*$ for the optimal policy and $Q^{*}$ and $V^*$ as its respective value functions. 
\iffalse

%\gls{DRL} problem can be formulated as a \gls{MDP}, which includes the state space $\mathcal{S}$, action space $\mathcal{A}$, transition probability function $\mathcal{P}$, reward function $\mathcal{R}$ and discount factor $\gamma$ in detail.

In terms of the target policy $\pi:\mathcal{S}\rightarrow\mathcal{A}$, a value function $V^\pi$ is designed as a description of total discounted reward $G_t$ for each state $s\in\mathcal{S}$:
\begin{equation}
    V^\pi(s) = \mathbb{E}_\pi[G_t|s_t=s]
\end{equation}
where $t$ is the discrete time step and $G_t$ is the expected return after time step $t$.

% \xiaowei{$t$ is not defined. the left hand side $V^\pi(s)$ has no $t$ but the right hand side has, why? }

With the Bellman equation \cite{bellman1966dynamic}, $V^\pi$ can be represented by a recursive form:
\begin{equation}
    V^\pi(s) = \mathbb{E}_\pi[r_t+\gamma V^\pi(s_{t+1})|s_t=s]
\end{equation}

The action-value function $Q^\pi$ is formulated as follows:
\begin{equation}
    Q^\pi(s,a) = \mathbb{E}_\pi[r_t+\gamma Q^\pi(s_{t+1},a_{t+1})|s_t=s,a_t = a]
\end{equation}
The \gls{DRL} algorithms are trying to find an optimal action that maximises the action-value function or the value function: $\pi^*(s) = \arg\max_{a\in\mathcal{A}} Q^*(s,a)$, here $Q^*(s,a)$ is the value of taking action $a$ in state $s$ under the optimal policy $\pi^*$. 
% \xiaowei{should be $\pi^*(s)$? also, $Q^*$ is not defined. }

\fi

\subsection{Probabilistic Model Checking}

\gls{PMC} \cite{kwiatkowska_probabilistic_2018} has been used to analyse quantitative properties of systems across a variety of application domains, including \gls{RAS} \cite{zhao_probabilistic_2019,zhao_towards_2019,gerasimou_undersea_2017}. It involves the construction of a probabilistic model, e.g.,  \gls{DTMC} or  \gls{MDP}, that formally represents the behaviour of a system over time. The properties of interest are usually specified with, e.g., \gls{LTL} or \gls{PCTL}. Then, via model checkers, a systematic exploration and analysis are performed to check if a claimed property holds. In this paper, we adopt \gls{DTMC} and \gls{PCTL} whose definitions are as follows. 

\begin{definition}[\gls{DTMC}]
Let $AP$ be a set of atomic propositions. A \gls{DTMC} is a tuple $(S,s_0,\textbf{P},L)$, where 
%\begin{itemize}
%\item 
$S$ is a (finite) set of states, $s_0\in S$ is an initial state, 
%\item 
$\textbf{P}:S\times S \rightarrow [0,1]$ is a probabilistic transition matrix such that $\sum_{s^{\prime}\in S}\textbf{P}(s,s^\prime)=1$ for all $s\in S$, and 
%\item 
$L:S\rightarrow 2^{AP}$ is a labelling function assigning  each state with a set of atomic propositions.
%from $AP$.
%\end{itemize}
\end{definition}
\begin{definition}[\gls{DTMC} Reward Structure]
A reward structure for DTMC $D=(S,s_0,\textbf{P},L)$ is a tuple $r=(r_S, r_T)$ where $r_S:S\rightarrow \mathbb{R}_{\ge 0}$ is a state reward function and $r_T:S\times S \rightarrow \mathbb{R}_{\ge 0}$ is a transition reward function.
\end{definition}

\begin{definition}[\gls{PCTL}]
The syntax of \gls{PCTL} is defined by \textit{state formulae} $\phi$, \textit{path formulae} $\psi$ and \textit{reward formulae} $\mu$.
\begin{align}
\phi &::= true \mid ap \mid \phi \wedge \phi \mid \neg \phi \mid {P}_{\bowtie p}(\psi) \mid {R}_{\bowtie q}^{r}(\mu)\nonumber
\\
\psi &::= \next \: \phi \mid 
%\phi \: U^{\leq t} \: \phi \mid 
\phi \: U \: \phi \nonumber  
\\
\mu &::= 
%I^{=t} \mid
C^{\leq t} \mid \eventually \: \phi \nonumber
\end{align}
where $ap \in AP, p\in [0,1], q\!\in \!\mathbb{R}_{\geq 0}, t\!\in\! \mathbb{N}$, $\bowtie \in \{<,\leq,>,\geq\}$ and $r$ is a reward structure.

The temporal operator $\next$ is called ``next'', and  %$U^{\leq t}$ is called ``bounded until'' while 
$U$ is called ``until''. 
We write $\eventually \, \phi$ 
%as a syntax sugar 
for $true \, U \, \phi $, and call it ``eventually''. Operator 
%$I^{=t}$ represents the state reward at time step $t$---``instantaneous reward'', while 
$C^{\leq t}$ is ``bounded cumulative reward'', expressing the reward accumulated over $t$ steps.
%(when $t=\infty$, we write $C$ for short as the total reward accumulated indefinitely). 
Formula ${R}_{\bowtie q}^{r}(\eventually \, \phi)$ expresses ``reachability reward'', the reward accumulated up until the first time a state satisfying $\phi$. %\xiaowei{two $F$ formulas. to diamond}

%State formula $\phi$ is evaluated 
%to be either true or false in 
%on states.
%each state. 
Given $D=(S,s_0,\textbf{P},L)$ and $r=(r_S, r_T)$, the satisfaction of state formula $\phi$ on a state $s\in S$ is defined as:
\begin{align}
s & \models true; \quad
s \models ap \, \Leftrightarrow\, ap \in L(s);\quad
s  \models \neg \phi \,\Leftrightarrow\, s \not\models \phi ; \nonumber
\\
s & \models \phi_1\wedge\phi_2 \, \Leftrightarrow\, s \models \phi_1 \text{ and } s \models \phi_2; \nonumber
\\
s & \models \mathcal{P}_{\bowtie p}(\psi) \,\Leftrightarrow\, Pr(s\models \psi)\bowtie p ; \nonumber
\\
s & \models \mathcal{R}_{\bowtie q}^{r}(\mu) \,\Leftrightarrow\, \mathbb{E}[rew^{r}(\mu)] \bowtie q, \nonumber
\end{align}
where $Pr(s\models \psi)\bowtie p $ concerns the probability of the set of paths satisfying $\psi$ starting in $s$. Given a path $\eta$, if write $\eta[i]$ for its \textit{i}-th state and $\eta[0]$ the initial state, then
{\small
\begin{align}
%rew^{r}(I^{=t})(\eta)&=r_S(\eta[t]) \nonumber
%\\
rew^{r}(C^{\leq t})(\eta)&= \sum_{j=0}^{k-1}(r_S(\eta[j])+r_T(\eta[j],\eta[j+1])) \nonumber
\\
rew^{r}(\eventually\phi)(\eta)&=\begin{cases} \infty & \forall j \in \mathbb{N}(\eta[j] \not\models \phi) \\ rew^{r}(C^{\leq ind(\eta,\phi)})(\eta) & \text{otherwise}\end{cases} \nonumber
\end{align}}\normalsize
where $ind(\eta,\phi)=\min\{j|\eta[j] \models \phi\}$ denotes the index of the first occurrence of $\phi$ on path $\eta$.
Moreover, the satisfaction relations for a path formula $\psi$ on a path $\eta$ is defined as:
\begin{align}
\eta & \models \next\phi \,\Leftrightarrow\, \eta[1] \models \phi \nonumber
\\
%\eta & \models \phi_1 \, U^{\leq t}\,\phi_2 \,\Leftrightarrow\,  \exists 0 \leq j \leq t \nonumber
%\\
% & \quad(\eta[j]\models \phi_2\wedge(\forall 0\leq k<j \; \eta[k]\models \phi_1))\\
\eta & \models \phi_1 \, U\,\phi_2 \,\Leftrightarrow\,  \exists j \geq 0 (\eta[j]\models \phi_2\wedge \forall k<j(\eta[k]\models \phi_1)) \nonumber
\end{align}
\end{definition}

Very often, it is of interest to know the actual probability that a path formula is satisfied, rather than just whether or not the probability meets a required threshold since this can provide a notion of margins as well as benchmarks for comparisons following later updates. Subsequently, the \gls{PCTL} definition can be extended to allow \textit{numerical queries} of the form $\mathcal{P}_{=?}(\psi)$ or $\mathcal{R}^r_{=?}(\psi)$ \cite{kwiatkowska_probabilistic_2018}.
After formalising the system behaviours and properties in \gls{DTMC} and \gls{PCTL}, respectively, 
%the verification focuses on checking of \textit{reachability} in a \gls{DTMC}. 
%That is, \gls{PCTL} expresses the constraints that must be satisfied, concerning the probability of, starting from the initial state, reaching some states labelled as, e.g., unsafe and success 
automated tools have been developed to solve the verification problem, e.g., PRISM \cite{kwiatkowska_prism_2011} and STORM \cite{dehnert_storm_2017}.

\section{Problem Formulation}\label{sec.4}
\subsection{Running Example}
We consider a DRL-driven robot that navigates, and avoids collisions, in a complex environment where there are static and dynamic objects/obstacles.  
%, due to its unlikely to learn models on all possible environments, and 
%in some cases it 
%with such a complex environment, the autonomous robot 
%is impractical to be trained in a real-world environment where the negative examples are of high costs. 
%with no priority knowledge 
%due to the high cost of negative examples. 
The model-free \gls{DDPG} algorithm \cite{lillicrap2015continuous} is applied for the training of a DRL policy for the robot. Typically, the DRL policies are trained in a simulation environment before being applied to the real world \cite{christiano2016transfer}. That is because of the unbearable costs of having real-world (negative) examples for training in the real world \cite{yu2018towards}. 

%Moreover, the studied autonomous robot system has continuous action sets and continuous state sets, which are hard to formulated with a discrete algorithm. Thus, a model-free, off-policy algorithm, named \gls{DDPG}, is applied in this paper. 

%To compatible with the \gls{DDPG} algorithm, 
As stated in Section~\ref{sec:mdp}, 
the robot 
%autonomous 
%system 
%has been 
can be modelled as an \gls{MDP}. At each time $t$, %the 
%autonomous 
%robot 
it has its observation of the laser sensors from the environment, namely state $s_t$, i.e.,  
%set as:
	\begin{equation}\label{states}
		s_t = (o^1_t,o^2_t,\cdots,o^n_t)^T
	\end{equation}
where 
%$s_t\in \mathcal{S}$ presents the observable information and 
$o^1_t,o^2_t,\cdots,o^n_t$ are 
%detailed 
sensor signals at time $t$. 
%$\mathcal{S}$ is state-space or all possible states of the robot in the environment. 
As usual, the sensors can only scan the environment within a certain distance, for example, it is within 3.15 metres in Turtlebot Waffle Pi \cite{name} for a distance sensor. 

%The 
An action $a_t\in\mathcal{A}$ consists of several decision variables.
%made by actor networks. 
With the \gls{PID} controller on the 
%autonomous 
robot, %the actor networks 
%it only needs to decides 
we consider two  action variables, representing line velocity and angle velocity, respectively, i.e., 
$a_t = (v^{line}_t, v^{angle}_t)^T$.
%where $v^{line}_t, v^{angle}_t$ are the line velocity and angle velocity, respectively. Here,  $\mathcal{A}$ is the set of all possible moves that the robot can make. 
At each time $t$, the DRL actor network outputs an action pair $(v^{line}_t, v^{angle}_t)$ from the action set $\mathcal{A}$.

The objective of the robot is to avoid obstacles and reach a goal area. On every state $s_t$, the sensory input $o^i_t$ can be utilised to, e.g., predict the distance to the obstacles and the goal area when they are close enough (within 3.15 metres). The environment imposes a reward function $r$ on both the states (w.r.t.  the distance to obstacles) and the actions (w.r.t. the acceleration in linear or angular speed). 

We leave out the details of training a DDPG agent, and only refer to the trained policy $\pi: {\cal S} \rightarrow {\cal A}$.

\begin{figure}[htbp]
\centerline{\includegraphics[width=\hsize]{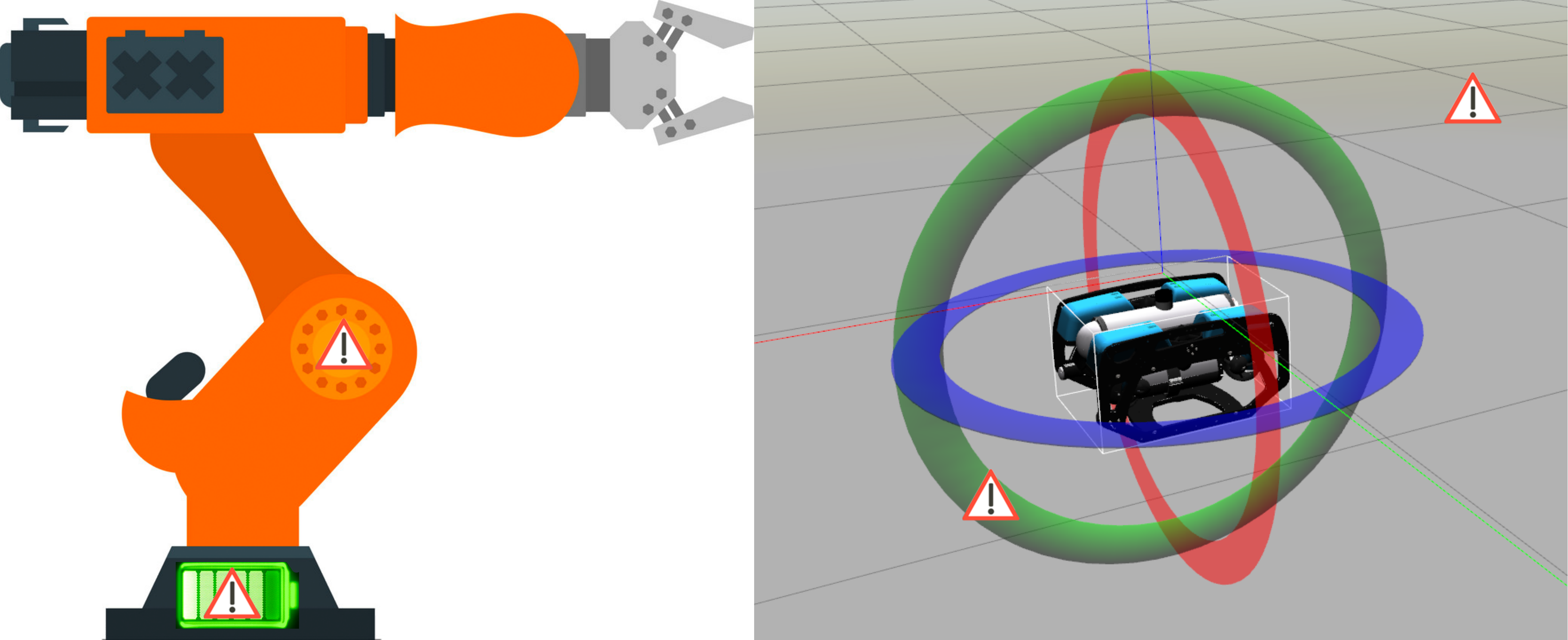}}
%\vspace{-0.5em}
\caption{Definitions on risk in different applications. LHS: mechanical failures and power supply shortage are risky situations for industrial robots. RHS: over-limited rotation angle and high liquid flow rates are risky situations for underwater vehicles.}
%\vspace{-6mm}}
\label{risks}
\end{figure}

\begin{remark}[Risk]
In different application contexts, ``risky situations" may vary case by case and typically are defined based on safety analysis like hazard identification for the given application.
While two examples are shown in Fig.~\ref{risks}, we define risk concerning the distance between the robot and the closest obstacle in this paper.
%as the different risk levels, and the collision/crash is treated as the catastrophic failure.
%1) mechanical failures and power supply shortage are the risky situations for industrial robots; 2) over-limited rotation angle and high liquid flow rates are the risky situations for underwater vehicles. In the running example, we use the distance between the robot and the closest obstacle as the different risk levels, and the collision/crash is treated as the catastrophic failure.
\end{remark}

\begin{remark}[Disturbance]
\label{rm_disturbance}
In real-world RAS applications, the environments of the robots are subject to different levels of disturbances due to, e.g., different wind speeds, weather conditions, and ground surfaces \cite{yang2018active, seo2019robust}. 
In this paper, the sensor noise is deemed to be the disturbances of the unmanned ground vehicle.
\end{remark}

\subsection{Construction of a DTMC Describing the Failure Process}\label{sec:DTMCconstruction}
%\xingyu{just in case we forgot.. I feel we probably want to mention our method is applicable to ``model-free DRL'' somewhere in this paper..}
%\xingyu{Sven and Nicolas' feedback of why even bother with DTMC if we have those sampled trajectories? Potential answer: (i) DTMC is a formal way of ``preprocessing'' the statistical data based on which automatic calculation tools can be invoked to quantify those properties. (ii) the DTMC may contain knowledge sourced from say experts rather than just statistical data.}

%\xingyu{Sven and Nicolas' feedback on if we lost important information because of the Markov property.. Potential answer: Yes, we lost some temporal information, but...}

\begin{figure}[htbp]
\centerline{\includegraphics[width=0.75\hsize]{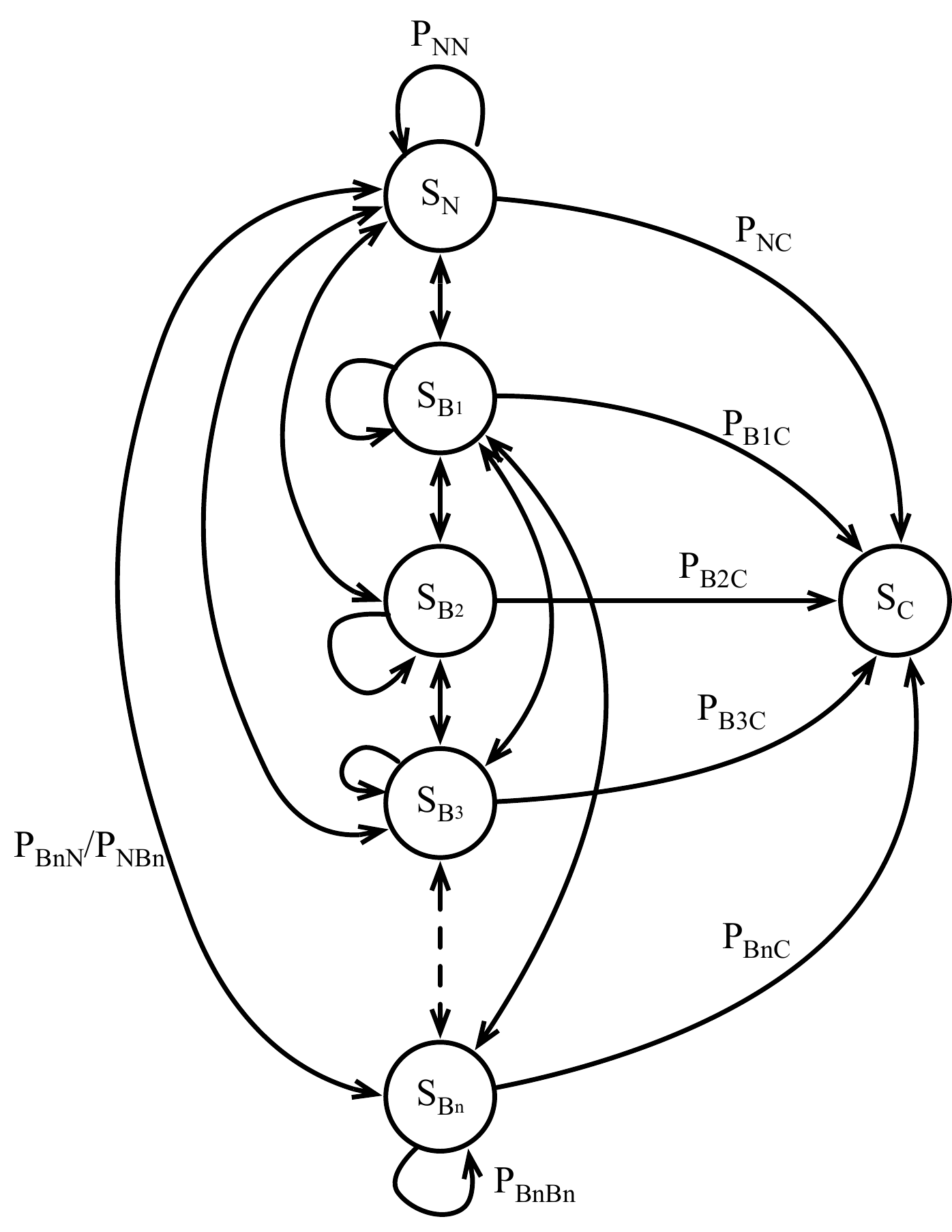}}
\centering
\caption{The failure process DTMC based on risk levels.}
\label{dtmc}
\end{figure}

We consider the execution of the policy $\pi$ in an environment. For simplicity, we only differentiate the environments with a disturbance level that the robot's sensory input may be subject to, and assume that the disturbance level follows a distribution $\mathcal{N}(0,\sigma)$.
Now, as stated in Section~\ref{sec:mdp}, given an MDP ${\cal M}^\sigma$ (based on a disturbance $\mathcal{N}(0,\sigma)$) and a DRL policy $\pi$, there is a trajectory distribution $\rho^{\pi,\sigma}(\zeta)$.  
Based on the \textit{dynamics of risk levels} in $\rho^{\pi,\sigma}(\zeta)$,  
%during the \gls{RAS} mission, 
we 
%first 
can define a DTMC, 
%structure 
%for the proposed assessment framework, 
as shown in Fig. \ref{dtmc}. It consists of a ``negligible-risk'' state $s_N$, a catastrophic failure state $s_C$, and several states $s_{B_i}$ representing different levels of ``benign failures''.

%(risky states that might lead to a catastrophic failure later). 
%TODO: remove all \vspace once accepted...

%Given a trained
%/frozen 
%\gls{DRL} policy, statistical testing is conducted to test the \gls{RAS} with different disturbance-levels (representing noisy environmental factors), yielding a set of \textit{mission trajectories}. Note, each
%We can use statistical testing to get a finite set of trajectories from $\rho^{\pi,\sigma}(\zeta)$. 
Each trajectory is a sequence of successive states %(or path) 
from the initial state to the end state of a DRL episode. First, we map each state in the trajectories to one of the states describing the failure process (i.e., $s_N$, $s_C$, and $s_{B_i}$). Second, we may conduct statistical analysis on the frequency of transitions between $s_N$, $s_C$, and $s_{B_i}$, based on which we 
%invoke estimators to 
estimate their corresponding transition probabilities. Finally, we construct the failure process DTMC with the defined structure and the estimated transition probabilities.
%(which is encoded later by some formal language and fed into model checkers). 
To be exact, we describe the 3 main steps above as what follows.
%The \gls{DRL} policy is abstracted into the induced DTMC, and the normal and safe states are modelled in the non-risky route state. In the proposed DTMC framework, each state can transit into one or more unsafe states with different probabilities, e.g. in a safe driving state $s_G$, the autonomous robot may fall into an unsafe state $B_3$ if there is a suddenly injected obstacle. We divide the unsafe states into several different levels, $B_1, B_2, \cdots, B_n$ to reveal different failure levels. The far from the safe states, the higher the possibility of mission failure, but none of the benign failures will cause any catastrophic damage to the autonomous robot.

\subsubsection{Mapping MDP States onto DTMC States}

% 
%Without loss of generality, 
%in this paper, 
First of all, every state in the DTMC (cf. Fig.~\ref{dtmc}) is associated with a risk level. Specifically, $s_N$ is the negligible-risk state, $s_C$ is the catastrophic
failure state, and $s_{B_i}$ are benign failure states such that the risk on $s_{B_i}$ is higher than on $s_{B_j}$ if $i>j$. 

Now, to map $\mathcal{S}$ (the states on the trajectories) onto $S$ (the states on the DTMC), we define a measure of risk based on the distance of the robot to obstacles. 
For instance, $s_N$ suggests that the robot is 3+ metres away from the obstacle, $s_{B_1}$ suggests 2-3 metres away, 
$s_{B_2}$ suggests 1-2 metres away, etc.  %and  
%but within 3m 
%represents state  
%(similarly for $s_{B2}$, etc), 
%and.
Moreover, catastrophic failure $s_C$ is defined as the robot terminated unexpectedly by a non-recoverable failure.  The determination of the risk levels for states in $\mathcal{S}$ can be done by evaluating the sensory input. 

%for every state $s\in \mathcal{S}$, we have a measure of risk. W.l.o.g.,
%our measure of risks
%we define the measure based on the distance to obstacles. 

\iffalse

\begin{definition}[Negligible-Risk State]
A state $s$ is mapped onto the negligible-risk state $s_N$  if the measure of risk on $s$ is below a pre-specified safe threshold.
%is defined as an .
\end{definition}
\begin{definition}[Benign Failure State]
A state, denoted as $s_{B_i}$, in which the measure of risk is above the safe threshold but lower than a specified level $B_i$ (while the \gls{RAS} mission is still ongoing) is defined as a benign failure state.
\end{definition}
\begin{definition}[Catastrophic Failure State]
A state, denoted as $s_{C}$, in which we observe the \gls{RAS} mission is terminated unexpectedly by a non-recoverable catastrophic failure is defined as a catastrophic failure state.
\end{definition}
\fi

%\noindent 
\begin{definition}[Negligible-Risk Route]
Given an MDP ${\cal M}^\sigma$ and a DRL policy $\pi$, a \textit{negligible-risk route} is defined as a mission trajectory in $\rho^{\pi,\sigma}(\zeta)$ that contains only $s_N$ states.
%the non-risky route in a \gls{RAS} environment. The \textit{Non-Risky Route} is the safest route, but it may not be the route with the highest reward under the concept of \gls{DRL}. \xiaowei{I believe we need a formal definition on the ``non-risky route''. Given a policy, there might not exist a path which stays on $s_G$. }
\end{definition}

 We remark that,
\iffalse
: (i) Although in some extreme 
cases 
%environments
with high-level of disturbance the negligible-risk route may not exist in practice, there is always a negligible-risk route in theory (potentially with 
%extreme 
small probabilities); (ii) The 
\fi
the negligible-risk route is not necessarily the optimal route achieving the highest reward,
%in the training of the DRL, 
rather it 
only depends on the 
%observations of the 
risk levels during the RAS mission.

\subsubsection{Estimating Transition Probabilities}

We can collect a set of mission trajectories by conducting 
%After the 
statistical testing (the simple Monte Carlo sampling in our case) on $\rho^{\pi,\sigma}(\zeta)$. 
%Given a trajectory, mapping each of its state to one of the states describing the failure process will result a sequence of states consisted of $s_N$, $s_C$ and $s_{B_i}$, and thus transitions among them as well. 
Then, all mission trajectories collectively can be transformed into a 
%large 
set of transitions, based on which we build a transition matrix to record the statistical data as follows:
\begin{table}[h!]
\centering
\begin{tabular}{l|lllll}
         & $s_N$       & $s_{B_1}$  & ... & $s_{B_m}$      & $s_C$         \\ \hline
$s_N$    & $n_{1,1}$   & $n_{1,2}$ & ... & $n_{1,m+1}$   & $n_{1,m+2}$   \\
$s_{B_1}$ & $n_{2,1}$   & $n_{2,2}$ & ... & ...           & ...           \\
...      & ...         & ...       & ... & ...           & ...           \\
$s_{B_m}$ & $n_{m+1,1}$ & ...       & ... & $n_{m+1,m+1}$ & ...           \\
$s_C$    & $n_{m+2,1}$ & ...       & ... & ...           & $n_{m+2,m+2}$
\end{tabular}
\end{table}
\newline
where $n_{1,1}$ records the number of transitions from $s_N$ to $s_N$, and so on.
%so forth. 
%While 
$m$ is the number of levels of benign failures (that varies case by case depending on the application-specific context, e.g., we choose $m=3$ in our 
%later 
experiments).

%Let us denote 
Let the transition probability matrix of the failure process DTMC be $\textbf{P}_1=(p_{ij})\in [0,1]^{(m+2)\times (m+2)}$. 
In a DTMC, given a current state $i$, the transition to the next state follows a \textit{categorical distribution}. Due to the Markov property, the categorical distributions of each state are \textit{independent}. Hence, as we observe repeated outgoing transitions from state $i$, the repeated categorical process follows a \textit{multinomial distribution}. For the $i$-th row of $\textbf{P}_1$, the likelihood function $\mathcal{L}$ is (by omitting the combinatorial factor):
\begin{equation}
\label{eq_likelihood_row_i}
\mathcal{L}(  p_{i,1},\dots,p_{i,m+2} \mid n_{1,1},\dots,n_{1,m+2})=\prod_{j=1}^{m+2} p_{i,j}^{n_{i,j}}
\end{equation}
%\xiaowei{I don't quite understand this likelihood function. }

Upon establishing the likelihood function, many existing estimators can be invoked for our purpose---from the basic Maximum Likelihood Estimation (MLE), Bayesian estimators \cite{epifani_model_2009} and estimator with (frequentist/Bayesian) bounds \cite{calinescu_formal_2016,zhao_probabilistic_2019}. While more advanced estimators can be easily integrated in our proposed framework, we only present the use of MLE in this paper for brevity:
\begin{equation}
\label{eq_mle_tranprob}
    \hat{p}_{i,j}=\frac{n_{i,j}}{\sum_{j=1}^{m+2} n_{i,j}}
\end{equation}
It is known that MLE is an unbiased estimator 
%in this case 
\cite{epifani_model_2009}, while the uncertainty in the estimates is captured by the variance that depends on the number of samples. The propositions in \cite{calinescu_formal_2016} provide the means for calculating $(1-\alpha)$ confidence intervals of the verification results, given the observations on the frequencies between states (exactly as our statistical data $n_{i,j}$). Such result may in turn determine the required number of samples $n_{i,j}$ given a required say 95\% confidence level for the final verification results. Although we did not calculate the confidence interval to determine the sample size in this paper, we instead chose a sample size in our later experiments that is sufficiently large to show a converging trend of the verification results (cf. Section~\ref{sec.5.3}). 

\subsubsection{Construction of Failure Process DTMC}
\label{sec_formalise_DTMC}
%Let $AP=\{term, p_G, p_{B^1},..., p_{B^n}\}$. From a trajectory distribution $\rho^\pi(\zeta)$ induced from a policy $\pi$ and an MDP $\cal{M}$,  we  construct a DTMC $(S,s_G,\textbf{P},L)$ such that $S=\{s_G, s_{B^1},...,s_{B^n}, s_C\}$,  $term \in L(s_C)$, and $p_{B^i}\in L{s_{B^i}}$ for $i\in \{1..n\}$. The probabilistic transition $\textbf{P}$ is defined as $\textbf{P}(s_G,s_1)=$
%\xingyu{double check with Xiaowei...}

The failure process DTMC is the product of two DTMCs, $M_1$ and $M_2$, via the synchronisation of the transition actions.

Let $AP_1=\{crash, neg\_risk, risk\_B_1,\cdots, risk\_B_n\}$, and 
%From a trajectory distribution $\rho^\pi(\zeta)$ induced from a policy $\pi$ and an MDP $\cal{M}$, 
we construct the first DTMC $M_1=(S,s_N,\textbf{P}_1,L_1)$ where $S=\{s_N, s_{B_1},\dots,s_{B_n}, s_C\}$,  $neg\_risk \in L_1(s_N)$, $risk\_B_i\in L_1(s_{B_i})$ for $i\in \{1..n\}$, and $crash \in L_1(s_C)$. Each entry $p_{i,j}$ of 
%the probabilistic transition matrix 
$\textbf{P}_1$ is defined as Eqn.~\eqref{eq_mle_tranprob}. We also define a reward structure 
%for this DTMC 
%as
{``deviation''}$=(r_S,r_T)$ with $r_S(s_N)=0$, $r_S(s_C)=0$, $r_S(s_{B_i})=d_i$ (where $d_i$ is the deviation from $s_N$ to $s_{B_i}$) and $r_T(s1,s2)=0$ for all $s1, s2 \in S$.

%Since the properties 
%of our interest 
%may depend on %in which 
%the stage the \gls{RAS} is during the mission, 
Moreover, we %also 
%formalise 
need a ``mission stage DTMC'' (for simplicity, we only consider two stages---mission terminated or not). Let $AP=\{progressing,terminated\}$, we construct
%a DTMC 
$M_2=(K,k_0,\textbf{P}_2,L_2)$ with $K=\{k_0,k_1\}$, $progressing \in L_2(k_0)$ and $terminated \in L_2(k_1)$. The transition probabilities $\textbf{P}_2$ are $p_{k_0,k_1}=\frac{1}{l_{mis}},p_{k_0,k_0}=1-\frac{1}{l_{mis}}, p_{k_1,k_1}=1$ and $p_{k_1,k_0}=0$, where $l_{mis}$ is a constant representing the expected mission length (number of transitions) obtained from the testing data. We also define a reward structure for this DTMC:  {``step''}$=(r_S,r_T)$ with $r_T(k_0,k_0)=1$, $r_T(k_0,k_1)=1$ and $r_S(k)=0$ for all $k \in K$.

%In later experiments, 
Finally,  
%further 
%encode the two DTMCs in the 
%for model checking, 
we encode the failure process DTMC with PRISM model checker \cite{kwiatkowska_prism_2011}. Although we only present how to leverage sampling to construct the DTMCs as an initial framework, other works have shown how to incorporate diverse evidence for which we have the following remark.
%\footnote{https://www.prismmodelchecker.org/manual/}.
%and do model checking based on  with .

%\xiaowei{We remark that, the above construction of DTMC is generic, and can be applied to all RAS applications with problem specific definitions on the mapping of MDP states onto DTMC states. The mapping actually defines the risk levels for the states, and can be obtained through e.g., hazard analysis. For example, ... }

\begin{remark}[Constructing DTMCs from Disparate Evidence]
\label{rm_dtmc_benefit}
In addition to statistical sampling, the DTMCs constructed in our method may contain knowledge sourced from disparate evidence, e.g., domain knowledge, \gls{vnv} and traditional safety/hazard analysis (say from FTA/FMEA to DTMCs) \cite{calinescu_efficient_2021,zhao_probabilistic_2019}. Such extra knowledge is formally incorporated via our failure process DTMC to accurately model the dynamics of risk in the executions of RAS.
\end{remark}

\section{Formal Properties}
\label{sec.5}

\newcommand{\misscomp}{miss\_comp}
\newcommand{\critsitu}{crit\_situ}
\newcommand{\noncritsitu}{ncrit\_situ}

%As a holistic assessment framework, we 
We define a set of quantitative dependability properties, and use  
 \gls{PMC} to check if they hold on the DTMC.
%constructed earlier. 
In what follows, we go through each property  with both informal description (referring to Fig.~\ref{properties}) and PCTL definition.

Before proceeding, we define 
\begin{equation}
\begin{array}{rcl}
    \misscomp & := & \neg   crash \land terminated \\
   \critsitu  & := & risk\_B_{max} \land progressing\\
   \noncritsitu & := & neg\_risk \land progressing\\
\end{array}
\end{equation}
where $\misscomp$ denotes a successful completion of the mission (i.e., $terminated$ without $crash$), $\critsitu$ denotes the robot is in critical situation (i.e., at the most serious benign failure level $risk\_B_{max}$ but still $progressing$), and $\noncritsitu$ denotes the robot is in a sufficiently safe situation (i.e., negligible risk and $progressing$).  

%To fit the different level of \gls{DRL} states in the failure process \gls{DTMC}, we define a "non-risky route" in DRL assessment, which represents the ideal execution of the DRL missions. For example, the fastest, safest and optimal path would be the non-risky route in our autonomous robot application. Then, we could have definitions of the following several properties, which are demonstrated in Fig. \ref{properties}.

\textit{a)}: Safety 
%informally 
requires that ``something bad will never happen''. Thus, we quantify the RAS safety 
%in this work 
as the probability of never reaching the catastrophic failure state (representing, e.g., crashes) before the mission is terminated normally.
\begin{definition}[Safety]
The safety property $\textit{Prop}_{S}$ measures the probability that  the \gls{RAS}, starting in the initial state\footnote{We assume the RAS always initialises in the negligible-risk state $s_N$.}, successfully completes the mission. 
%will 
%never reach 
%to 
%the catastrophic failure state $s_C$ during its mission. 
Formally, 
%We formally define it as the probability of 
\begin{equation}
    \textit{Prop}_S := {P}_{=?}[\ \eventually\ \misscomp\ ]
\end{equation}
%where $(\neg   crash \land terminated)$ expresses that the robot terminates on a non-crash state. Therefore, quantitative safety concerns the probability of the robot terminating without failure. 
%\xiaowei{use regular logic expression}
%are variables representing the states of the two DTMCs defined in the PRISM model, cf. line 14 and 25 in Fig.~\ref{fig_prism_modules}. 
\end{definition}

%Safety: A safety predicate $\mathcal{M}_s$ comprises a initial state $s_C$, a set of unfailure states, and a step bound $k\in\mathbb{N}\leq\infty$. It states that the probability of safely reaching the target without catastrophic failures:where $K$ is the terminate signal and $s_C$ is the catastrophic state.
%Resilience is associated with a system, and it is an ability of a system to recover its function due to partial damage of the system  

\textit{b)}: Resilience, despite the existence of various definitions in the literature, is %normally 
generally referred as the ability 
%that a system 
to respond to change and 
%to 
survive/prosper,
%when challenged, 
e.g., %a system that can 
the ability to deal with attacks or surprised disturbances \cite{woods2017resilience}. %For  \gls{RAS}, we define the resilience from the perspective of the DRL policies. 
For RAS, we define it as an ability of the DRL policy that can help the \gls{RAS} recover from any deviations from the negligible-risk route. 
\begin{definition}[Resilience]
Given the reward structure 
%of 
``deviation'' (cf. Section \ref{sec_formalise_DTMC}), resilience is defined as the \textbf{expected total deviation} from the negligible-risk route in a successful\footnote{Presumably, there is no practical meaning to consider the resilience in a crashed mission.} mission. Formally, 
%that is:
%{\small
\begin{equation}
    \textit{Prop}'_{Res} := {R}^{\text{``deviation''}}_{=?}[\eventually \  \misscomp]
\end{equation}
%}\normalsize

To accord with the intuition, we normalise it 
%$\textit{Prop}'_{Res}$ 
%into $[0,1]$
by considering $max\_dev:=\max_i\{d_i\}\cdot  {R}^{\!\!\text{``step''}\!}_{=?}[\eventually \misscomp]$, the worst-case total deviation that could happen in a 
%single 
mission, which is the product of the largest deviation $\max_i\{d_i\}$ and the expected length of a mission that terminates safely, i.e., 
%and make higher values indicate better resilience:
%{\small
\begin{equation}
\label{eq_resiliecence_normlise}
\textit{Prop}_{Res} \!:=\!1-\frac{\textit{Prop}'_{Res}}{max\_dev}
\end{equation}
%}\normalsize
%where $max\_dev=\max_i\{d_i\} R\{\!\!\text{``step''}\!\}_{=?}[F(\neg   crash \land terminated)]$ is .
\end{definition}

\begin{comment}
To assess the resilience property, we define a return signal for every DTMC state. For example, if the agent falls into $B_1$ from Safe states, then return $R_1$. If the robot falls into state $C$, then the robot would be crashed with no return.
\begin{equation}
    S\rightarrow R_0, B_1\rightarrow R_1, B_2\rightarrow R_2, \cdots, B_n\rightarrow R_n.
\end{equation}
In our definition, resilience is, essentially, the area below the dotted line and above the dashed/solid line in Fig. \ref{properties}.Therefore, we could use return signals to approximate the expected offset area that away from the negligible-risk route in a mission:
\begin{equation}
    \mathcal{M}_{Res} = \mathbb{E}[\int_0^kR_tdt]
\end{equation}
It could be transferred into the formal language form:
\begin{equation}
    \mathcal{M}_{Res} = R_{=\mathord{?}}[F\ (K=1)]
%  R_{=\mathord{?}}[F\ \ (s!=s_S) \& (s!=s_C)]
\end{equation}
where the returns of reaching $B_1$, $B_2$ and $B_3$ are defined as $R_1$, $R_2$ and $R_3$, such that $R_1 \leq R_2 \leq R_3$, representing the “distance” from those 3 levels of benign failures to the negligible-risk route.
\end{comment}

\textit{c)}: Robustness 
%is a challenge faced by all \gls{DNN} based methods. It 
requires that the behaviour of the RAS is invariant against small disturbance on inputs. Formally, %In this paper, we define 
\begin{definition}[Robustness]
Given a disturbance level, the robustness $\textit{Prop}_{Rob}$ 
%is about 
quantifies the ability to resist the disturbance, i.e., recovering from the critical situation.
%and recover. 
%the most ``far-away'' benign failure level---the state $s_{B_i}$ associated with the biggest deviation from the non-risky route, denoted as $s_{B_{\textit{max}}}$. 
%Formally, %it is a conditional probability:
%{\small
\begin{align}
\textit{Prop}_{Rob} \!:=\!\frac{\!{P}_{=?}[ \eventually \  (\critsitu \land \eventually \, \misscomp)]}{{P}_{=?}[ \eventually \ \critsitu]}
\end{align}%}\normalsize
%which intuitively says, given the path reaches %the 
%$s_{B_{\textit{max}}}$ during the mission, the probability that it will 
%also 
%terminate without crash.
\end{definition}

\begin{comment}
two types of robustness depend on how different levels of disturbance are applied on the \gls{RAS}, as what follows:
\begin{definition}[Robustness]
Given a constant disturbance level $dl_i$, the intra-disturbance robustness $\textit{Prop}^{dl_i}_{Rob1}$ is about the reachability to the most ``far-away'' benign failure level---the state $s_{B_i}$ associated with the biggest deviation from the negligible-risk route state $s_{G}$, denoted as $s_{B_{\textit{max}}}$. Formally, we define it as a probability (specified in \gls{LTL}-style):
%say $s_{B_{\textit{max}}}=\max\{s_{B_1},\dots,s_{B_n}\}$.
\begin{align}
&\textit{Prop}^{dl_i}_{Rob1}\!:=\nonumber
\\
&\!P_{=?}[F (s\!=\!s_{B_{\textit{max}}})\&(k=0) \& (F (s!=s_C)\&(k=1))]
\end{align}
\end{definition}
\begin{definition}[Inter-Disturbance Robustness]
Given two different disturbance levels $dl_i$ and $dl_j$, we define inter-disturbance robustness $\textit{Prop}^{dl_{ij}}_{Rob2}$ as the absolute difference between the two intra-disturbance robustness, that is:
\begin{equation}
\textit{Prop}^{dl_{ij}}_{Rob2}:=|\textit{Prop}^{dl_i}_{Rob1}-\textit{Prop}^{dl_j}_{Rob1}|
\end{equation}
\end{definition}
\end{comment}

\textit{d)}: Detection intuitively refers to the ability of 
%that how quick the \gls{RAS} can 
``realising'' the existence of the disturbance and then starting to recover. 
\begin{definition}[Detection]\label{def:detection}
The detection property $\textit{Prop}'_{D}$ is defined as the number of steps (i.e. transitions in the mission-stage DTMC) 
%between two time-epoch in the mission: 
between the times when the disturbance is first applied\footnote{For simplicity, we assume the disturbance level is applied at the very beginning of the mission and does not change thereafter.}
%, while this assumption can be relaxed to any time-epoch during the mission.} 
and when the \gls{RAS} first reaches the critical situation  % $s_{B_{\textit{max}}}$. Again, we utilise the rewards in \gls{PCTL} to formally define it as:
%{\small
\begin{equation}
   \textit{Prop}'_{D} := R^{\text{``step''}}_{=?}[ \eventually \  \critsitu]
\end{equation}%}\normalsize
where the reward structure 
%of 
``step'' is defined in Section \ref{sec_formalise_DTMC}. Again, we normalise it by letting  %$\textit{Prop}'_{D}$ to $0-1$ and make higher values means higher detection-ability:
\begin{equation}
\label{eq_detect_normalise}
\textit{Prop}_{D} :=1-\frac{\textit{Prop}'_{D} }{R^{\text{``step''}}_{=?}[ \eventually \  \misscomp]}
\end{equation}
where the denominator is the expected length of a mission that terminated safely.
\end{definition}

\begin{comment}
Detection of a DRL policy is defined as the number of states between adversarial start state and the state that reach the maximum $B_n$ level. \begin{equation}
    0\leq \mathcal{M}_{D}=\mathbb{E} [n_D] \leq\infty
\end{equation}
In other words, the detection property is formulated as an accumulated reward if we define reward = 1 for each transition representing ``number of steps":
\begin{equation}
    \mathcal{M}_{D} = R_{=\mathord{?}}[F\ ((s!=s_{B_n})\& (K=0)]
\end{equation}
\end{comment}

\textit{e)}: Recovery intuitively refers to the ability of 
%how quick the \gls{RAS} will go back 
``recovering'' from a critical situation and eventually reaching the negligible-risk state, after the detection of the disturbance.  
\begin{definition}[Recovery]
The recovery property $\textit{Prop}_{Rec}$ is defined as the number of steps between 
%two time-epochs: 
the times when the disturbance is first detected and when the \gls{RAS} first reaches the $s_N$ after the detection, i.e.,
%We formally define it as:
{\small
\begin{align}
   \textit{Prop}_{Rec}' := R^{\text{``step''}}_{=?}[\, \eventually ( \critsitu \land \eventually \, \noncritsitu)] -\textit{Prop}'_{D}
\end{align}}\normalsize
It can be normalised in a similar way as in Definition~\ref{def:detection} to get $\textit{Prop}_{Rec}$, and we omit it for brevity.
\end{definition}

\begin{remark}[Generalisability]
The set of formally defined dependability properties is generic and can be applied to all RAS applications with problem specific definitions on the risk that maps onto failure process DTMC states.
\end{remark}

\begin{comment}
is defined as the number of states between benign failure level $B_n$ and the new equilibrium $B_e$. If cannot recovery to an equilibrium level, then $\mathbb{E}[n_R] = \infty$. 
\begin{equation}
    \mathcal{M}_{Rec} = \mathbb{E}[n_R] \geq 0
\end{equation}
is the ability of the policy that can recovery from the benign failure level $B_n$. 
\end{comment}

\section{Experimental Results}\label{sec.6}

This section presents experimental results regarding  the following research questions: 

%\noindent (\textbf{RQ1}) Are dependability properties correlated with the number of sampled trajectories (used in constructing DTMC)? \xingyu{shall we say sensitivity analysis?}
\noindent \textbf{RQ1}: How sensitive are the verification results to the number of sampled trajectories (used in constructing the DTMC)? 

\noindent \textbf{RQ2}: Does the DDPG training improve its dependability? %and 

\noindent \textbf{RQ3}: How do the dependability properties react to different levels of disturbances? 

%Can we apply the dependability analysis to study e.g., the DRL's Sim-to-Real challenge? 

In the following, we will first introduce our experimental environment, and then address the \textbf{RQ}s individually\footnote{We omit the verification of the recovery property in the experiments, due to the limitation of state-of-the-art model checker (e.g., PRISM and STORM)---they cannot calculate rewards with nested \gls{LTL}.}.

%regarding different testing episode numbers, policies and environments via the proposed holistic assessment method. At first, the simulation environment setups are introduced in section \ref{sec.5.1}, as well as the platform of the studied robotic autonomous systems in \ref{sec.5.2}. Then three test cases are simulated to verify the effectiveness of the proposed holistic assessment method. The first test case in \ref{sec.5.3} is to assess the influence of different testing episode numbers. The second test case in \ref{sec.5.4} is to reveal the effects among different policies. Finally, the third test case in \ref{sec.5.5} is to investigate the influence of different environments.

\subsection{Experimental Environment}\label{sec.5.1}

We have both simulation and physical environments for our experiments\footnote{All source code, DRL models, datasets, PRISM files and experiment results are publicly available at our project website \url{https://github.com/YD-19/HAM4DRL.git}}. 
%The objective of the RAS is to reach the goal in a stochastic environment with several obstacles.
For the simulation environment, we use ROS \cite{quigley2009ros} and Gazebo \cite{koenig2004design}. The training of DDPG algorithm is conducted on a maze space with obstacles.

\begin{figure}[htbp]
\centerline{\includegraphics[width=\hsize]{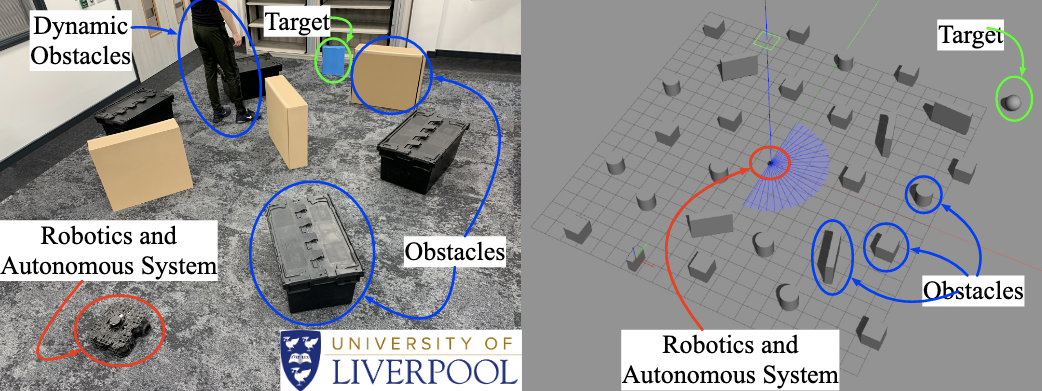}}
\captionsetup{justification=centering}
\caption{Experiment Environment\\ (Left: Physical; Right: Simulation).
%\vspace{-6mm}
}
\label{TurtlebotWafflePi}
\end{figure}

%Firstly, ROS \cite{quigley2009ros} and Gazebo \cite{koenig2004design} are applied as simulation environments to help train a collision avoidance DRL policy for the real robot experiment. The training process is carried out with designed DDPG algorithm, and a maze space with several obstacles. Then, the trained DRL policy is abstracted into a DTMC failure model based on Monte-Carlo algorithm. Finally, with some formally specified properties, different properties of the DRL algorithm can be assessed by the probabilistic model checking tools, such as PRISM \cite{kwiatkowska_prism_2011} and STORM \cite{dehnert2017storm}. %DRL
%PRISM
For the physical environment, we assemble a Turtlebot3 Waffle Pi as the robot, together with a laboratory environment where a number of static and dynamic objects are randomly placed,  
%which is 
as shown in Fig. \ref{TurtlebotWafflePi}. The Turtlebot3 is a compact and fully programmable mobile robot with a $360^o$ LiDAR onboard. It is based on the standard ROS platform.
%for SLAM and navigation purposes. 
The LiDAR information is taken as the input to the trained DRL policy. 
%used for DRL training and collision avoidance.
%\eqref{states}. 
 A Raspberry Pi and OpenCR are used to trigger two motors once received the output from the DRL policy. 
%has published the action information \eqref{actions}. 
%The robot is initialised at a random position with a random heading directions, in order to better simulate its performance in an unknown environment.

%\subsection{Robot Platform}\label{sec.5.2}

% \begin{figure*}[htbp]
%      \centering
%      \begin{subfigure}[b]{0.3\textwidth}
%          \centering
%          \includegraphics[width=\textwidth]{Figs/MCsafety.eps}
%          \caption{Different number of testing episode.}
%          \label{differenttesingepisodes}
%      \end{subfigure}
%      \hfill
%      \begin{subfigure}[b]{0.3\textwidth}
%          \centering
%          \includegraphics[width=\textwidth]{Figs/Psafety.eps}
%          \caption{Different number of training episode.}
%          \label{differentpolicies}
%      \end{subfigure}
%      \hfill
%      \begin{subfigure}[b]{0.3\textwidth}
%          \centering
%          \includegraphics[width=\textwidth]{Figs/SRsafety.eps}
%          \caption{Different running environments.}
%          \label{SRR}
%      \end{subfigure}
%         \caption{Simulation Results}
%         \label{fig:three graphs}
% \end{figure*}
\begin{figure*}[htbp]
\centerline{\includegraphics[width=\hsize]{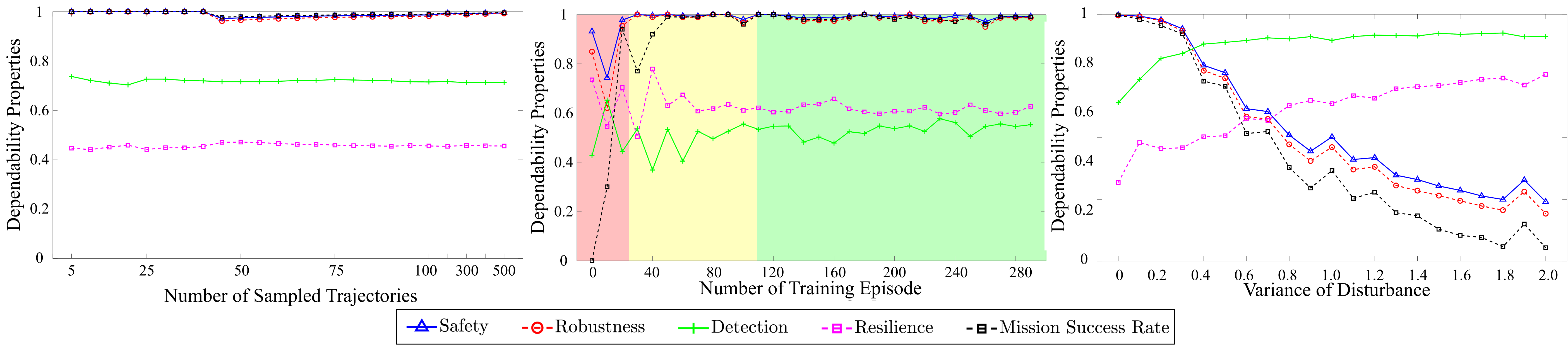}}
\captionsetup{justification=centering}
\caption{Dependability analysis with (a.) different number of Samples, \\(b.) different training episodes, and (c.) different disturbance levels.%\vspace{-6mm}
}
\label{resultsfig}
\end{figure*}%\vspace{-5mm}

\subsection{Dependability w.r.t. Sampled Trajectories}\label{sec.5.3}

To answer \textbf{RQ1}, we evaluate the dependability properties against the number $n$ of sampled trajectories. %For a given well-trained DRL policy, we test various properties in different testing episode numbers, 
We consider the cases where the number $n$ is gradually increased to 500. It is noticed that all the trajectories are sampled based on a given policy.
%from 100 to 1000. We do not consider the cases when $n<100$, because collecting 100+ trajectories is not only feasible in our experiments but also needed for a dependability analysis. \xingyu{TODO, rewrite based on Yi's new results of $n<100$}
% \begin{figure}[htbp]
% \centerline{\includegraphics[width=\hsize]{Figs/MCsafety.eps}}
% \caption{Properties with different testing episodes.}
% \label{differenttesingepisodes}
% \end{figure}
%
As given in Fig.~\ref{resultsfig}a, the properties (safety, robustness, detection, and resilience) are all stable w.r.t. $n$ in a typical experiment setting (a policy trained by some number of episodes and a certain disturbance level). We also repeat the experiment with other settings (used in later \textbf{RQ}s), and the overall results suggest that our dependability analysis in the next two \textbf{RQ}s are insensitive to the sample size especially when $n\geq300$ which is the number of samples used in later experiments.

%we can see that as the number of testing episodes increases, the property gradually stabilize. For example, after 300 testing episodes, the value of these properties do not change much. Therefore, the proposed assessment method does not require massive testing episode numbers and therefore can be applied in real complex environments.
\subsection{Dependability in Training}\label{sec.5.4}
%real-robotics
To answer \textbf{RQ2}, we apply the dependability analysis to monitor the changes of dependability properties during the training. We record the properties for the first 300 training episodes, as elucidated in Fig.~\ref{resultsfig}b. 
%In this section, we verify the effectiveness of the proposed holistic assessment framework in the training process. To better present the differences between those DRL policies, we record the properties in the first 300 training episodes, which are shown in Fig. \ref{differentpolicies}. 
During the training, the motion of the robot is terminated if and only if it crashes, reaches the goal, or gets stuck for 1000+ steps.

% \begin{figure}[htbp]
% \centerline{\includegraphics[width=\hsize]{Figs/TESTsafe.eps}}
% \caption{Properties with different training episodes.}
% \label{differentpolicies}
% \end{figure}

The values of the properties have significant fluctuations at the beginning of the training. This is due to the drastic weight update of DRL policies  by the training process. As the number of training episodes increases, these properties gradually stabilise. Due to the existence of the exploration process, the properties retain a certain level of fluctuations, until reaching the final convergence (with minor oscillation).
%, including safety, resilience, robustness, and detection. 
%These dependability properties are formulated under the definition of the multi-level DTMC failure models.
%Furthermore, 
Importantly, the training process 
%of the reinforcement learning 
can be roughly split into three phases: $conservative\ phase$ (red area in Fig.~\ref{resultsfig}b), $optimising\ phase$ (yellow area), and $optimised\ phase$ (green area). Due to the high cost of crashes, the DRL policy explores the environment safely and carefully 
%to reach its goal 
in the $conservative\ phase$, without reaching a goal. Therefore, the mission success rate at this stage is low, while the safety property remains high. Once the robot reaches its goal for the first time, it will receive a much higher reward, which in turn will influence all the state values in the path according to the training algorithm.
%based on the discounted factor. 
In the $optimising\ phase$, the DRL algorithm optimises the policy through exploration. Failed missions still exist  because the exploration may cause crashes. During this phase, the safety and robustness are kept high to ensure safe exploration, while other properties remain fluctuating.
%with the change of the parameters of DRL policy. 
In the $optimised\ phase$, the policies are already well-trained, and the properties are converged. The remaining small oscillation is due to environment disturbances.

%Therefore, the effectiveness of the proposed assessment algorithm is verified by the multi-level DTMC framework. Moreover, it can be found that the performances of the policies would not change much after several training episodes. 

Nevertheless, we notice that the training does not improve, only stabilise, properties such as resilience and detection. This is mainly because the training, as a stochastic optimisation process, does not incorporate optimisation objectives concerning these properties. This forms our future work. 
%\xingyu{Spell out what should be done?} 

%The DRL algorithms could be designed based on the performance requirements of those properties.
\subsection{Dependability w.r.t. Disturbance Levels}\label{sec.5.5}
%x:noise level;
%y:property level.
 %To verify whether the dependability analysis method can adapt different environments, a well-trained policy is selected and tested under a multi-levels disturbance circumstance.  In this case, 
As suggested in Section~\ref{sec:DTMCconstruction}, we 
 %model the environment with its disturbance level $\sigma$ and 
 use a half-Gaussian distribution 
 %disturbance 
$\mathcal{N}(0,\sigma)$ 
to model the disturbance level of the environment, and differentiate the environments with the parameter $\sigma$. This is without loss of generality, as we can extend this to more involved ways of modelling disturbances. 
 %The Gaussian distributions are zero mean and 
 In our experiments, $\sigma$ ranges from 0.1 to 2.0, and we take a well-trained policy $\pi$ for analysis. 
%\xingyu{Do we need to mention the absolute values of noises somewhere? Or say the folded normal distribution?}
 %At the mean time, it is assumed that the mission steps of the RAS system are 560 steps, which means that the proposed algorithm assesses various properties within 560 steps. However, we apply normalised detection and resilience properties to illustrate the performance of the trained policy. 

% \begin{figure}[htbp]
% \centerline{\includegraphics[width=\hsize]{Figs/SRsafety.eps}}
% \caption{Properties with different environments.}
% \label{SRR}
% \end{figure}

As shown in Fig.~\ref{resultsfig}c, we can compute the dependability properties of $\rho^{\pi,\sigma}$ by varying $\sigma$, so that the analysis is conducted over a set of environments with different levels of disturbances. 
%The robot is controlled by the DRL algorithm, which target is to reach the goal.  Therefore the robot will not continue to circle in a safe area. The blue line shows the property of safety, we 
We can see that, the RAS becomes less safe and robust (cf. the blue curve and the red curve, respectively) when the disturbance level increases. The trends are roughly aligned with the decrease of the mission success rate. 
%The red line represents the robustness.  %Due to the uncertainty of the disturbance, the system has a certain possibility of going in the worse direction, or may go in a better direction. Therefore, the robustness properties get worse at the begin, and then they will show a state of volatility when the disturbance is large to a certain extent.
% It shows that the robustness property gets worse at the begin, but when the disturbance continues to grow, the robustness of the RAS system would become better. This is because the disturbance in the environment does not always make the RAS system worse, but may also make the RAS system better.
%Green line illustrates the performance of the detection property. 
Also, it is unsurprising that the detection 
%speed of the RAS system 
becomes easier as the increases of disturbance level (cf. the green curve), because the robot may be easier, and faster, to reach a critical situation. 
%, which means that the disturbances existing in the environment makes the detection of the autonomous robot system more and more sensitive. 
%The purple line shows the resilience property of the DRL policy. We can see that 
On the other hand,
%it is somewhat surprising that, 
%as the disturbance increases, 
the resilience property shows a trend of getting better (cf. the purple curve). This may be related to the fact that, with greater disturbances, there are more risky routes, 1) many of which may be easily recovered and thus lead to the improvement to the resilience property; 2) most routes with worse resilience are crashed and thus excluded from the calculation by our definition of resilience.
% \xingyu{Yi to add/rewrite this part with the second reason} 
%If the disturbance becomes larger, the robot will be more likely to crash when approaching the goal. A better resilience policy should be applied to ensure the safe reach of the robot to the goal.
%We also find that there are some trade-offs between different properties, for example, the resilience and detection performance have opposite trends to the performance of safety and robustness. 

Notably, we observe two trends in Fig.~\ref{resultsfig}c---curves of detection and resilience increase while the curves of other properties decrease. This reveals the conflicts between different proprieties, which may require trade-offs in training.
\section{Conclusion 
%and Future Works
}\label{sec.7}
% \vspace{-1mm}
A dependability analysis framework  is proposed to evaluate a set of quantitative properties of a given DRL policy, such as safety, resilience, robustness, detection and recovery.
%while revealing some interesting trade-off features between properties. 
Experimental results show the effectiveness of the framework in assessing DRL-driven \gls{RAS} holistically.
%The proposed framework can test these properties within a well-designed multi-level DTMC failure model. The effectiveness of the proposed algorithm and inter-property balance are demonstrated by several simulation studies.
% \begin{table}[htbp]
% \caption{Table Type Styles}
% \begin{center}
% \begin{tabular}{|c|c|c|c|}
% \hline
% \textbf{Table}&\multicolumn{3}{|c|}{\textbf{Table Column Head}} \\
% \cline{2-4} 
% \textbf{Head} & \textbf{\textit{Table column subhead}}& \textbf{\textit{Subhead}}& \textbf{\textit{Subhead}} \\
% \hline
% copy& More table copy$^{\mathrm{a}}$& &  \\
% \hline
% \multicolumn{4}{l}{$^{\mathrm{a}}$Sample of a Table footnote.}
% \end{tabular}
% \label{tab1}
% \end{center}
% \end{table}

\section*{Acknowledgment \& Disclaimer}
\includegraphics[height=8pt]{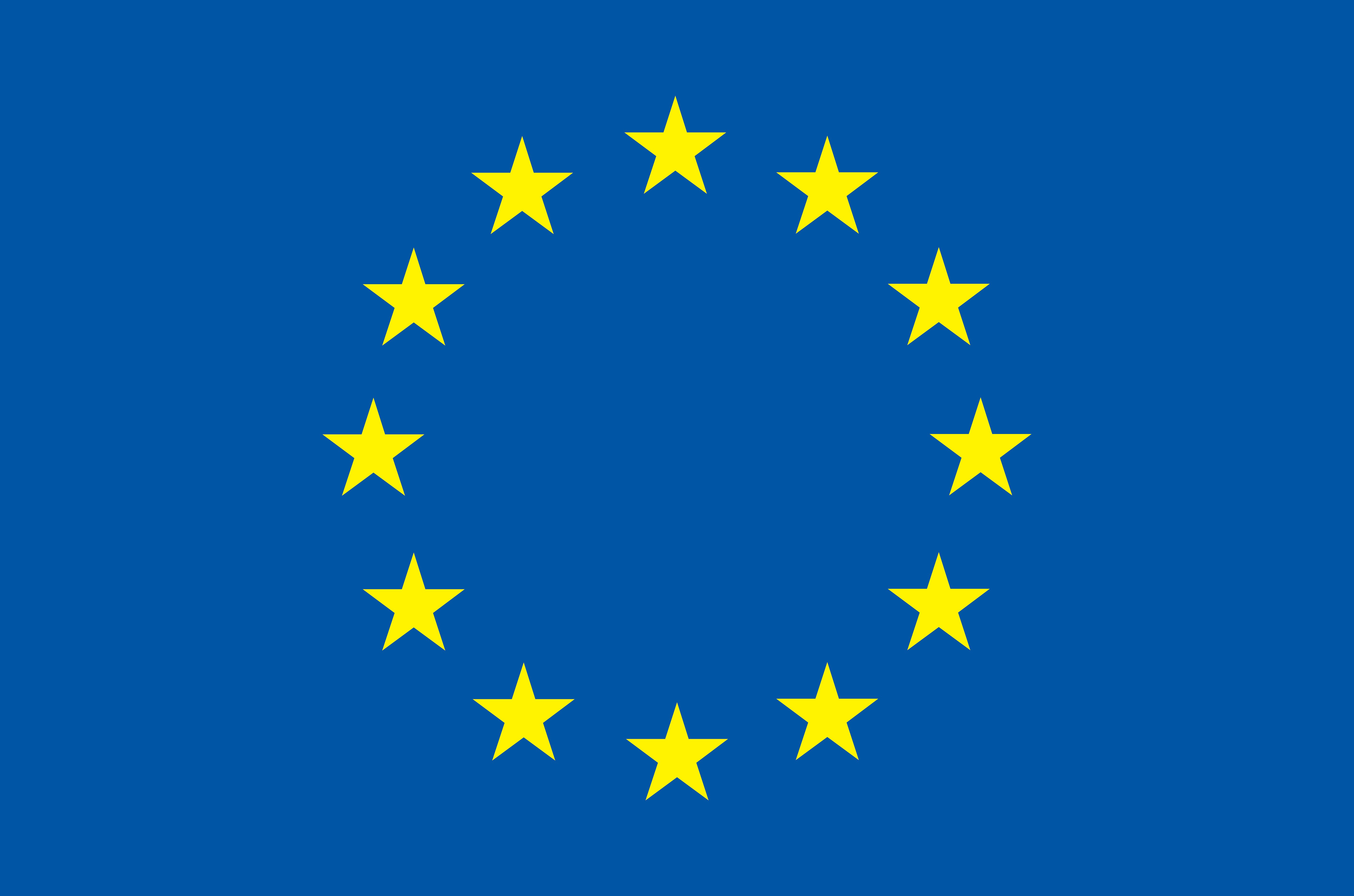} 
This project has received funding from the European Union’s Horizon 2020 research and innovation programme under grant agreement No 956123.
This work is supported by the UK Dstl through the project of Safety Argument for Learning-enabled Autonomous Underwater robots and the UK EPSRC through End-to-End Conceptual Guarding of Neural Architectures [EP/T026995/1].
XZ's contribution is partially supported through Fellowships at the Assuring Autonomy International Programme.

%XZ: without authors from DSTL, it seems we don't need this disclaimer---saving us some spaces :-)
%This article is an overview of UK MOD sponsored research and is released for informational purposes only. The contents of this article should not be interpreted as representing the views of the UK MOD, nor should it be assumed that they reflect any current or future UK MOD policy. The information contained in this article cannot supersede any statutory or contractual requirements or liabilities and is offered without prejudice or commitment.

%\clearpage

% \textcolor{red}{Do we need to add some references from IROS? }

\bibliographystyle{IEEEtran}
\bibliography{references}

\begin{thebibliography}{10}
\providecommand{\url}[1]{#1}
\csname url@samestyle\endcsname
\providecommand{\newblock}{\relax}
\providecommand{\bibinfo}[2]{#2}
\providecommand{\BIBentrySTDinterwordspacing}{\spaceskip=0pt\relax}
\providecommand{\BIBentryALTinterwordstretchfactor}{4}
\providecommand{\BIBentryALTinterwordspacing}{\spaceskip=\fontdimen2\font plus
\BIBentryALTinterwordstretchfactor\fontdimen3\font minus
  \fontdimen4\font\relax}
\providecommand{\BIBforeignlanguage}[2]{{%
\expandafter\ifx\csname l@#1\endcsname\relax
\typeout{** WARNING: IEEEtran.bst: No hyphenation pattern has been}%
\typeout{** loaded for the language `#1'. Using the pattern for}%
\typeout{** the default language instead.}%
\else
\language=\csname l@#1\endcsname
\fi
#2}}
\providecommand{\BIBdecl}{\relax}
\BIBdecl

\bibitem{lane_new_2016}
D.~Lane, D.~Bisset, R.~Buckingham, G.~Pegman, and T.~Prescott, ``New foresight
  review on robotics and autonomous systems,'' Lloyd’s Register Foundation,
  London, U.K., Tech. Rep. No. 2016.1, 2016.

\bibitem{avizienis_basic_2004}
A.~Avizienis, J.~Laprie, B.~Randell, and C.~Landwehr, ``Basic concepts and
  taxonomy of dependable and secure computing,'' \emph{IEEE Tran. on Dependable
  and Secure Computing}, vol.~1, no.~1, pp. 11--33, 2004.

\bibitem{mnih2013playing}
V.~Mnih, K.~Kavukcuoglu, D.~Silver, A.~Graves, I.~Antonoglou, D.~Wierstra, and
  M.~Riedmiller, ``Playing atari with deep reinforcement learning,''
  \emph{arXiv preprint arXiv:1312.5602}, 2013.

\bibitem{dong2021strategic}
Y.~Dong, Z.~Dong, T.~Zhao, and Z.~Ding, ``A strategic day-ahead bidding
  strategy and operation for battery energy storage system by reinforcement
  learning,'' \emph{Electric Power Systems Research}, vol. 196, p. 107229,
  2021.

\bibitem{sallab2017deep}
A.~E. Sallab, M.~Abdou, E.~Perot, and S.~Yogamani, ``Deep reinforcement
  learning framework for autonomous driving,'' \emph{Electronic Imaging}, vol.
  2017, no.~19, pp. 70--76, 2017.

\bibitem{cheng2019end}
R.~Cheng, G.~Orosz, R.~M. Murray, and J.~W. Burdick, ``End-to-end safe
  reinforcement learning through barrier functions for safety-critical
  continuous control tasks,'' in \emph{Proceedings of the AAAI Conf. on
  Artificial Intelligence}, vol.~33, no.~01, 2019, pp. 3387--3395.

\bibitem{hu2020voronoi}
J.~Hu, H.~Niu, J.~Carrasco, B.~Lennox, and F.~Arvin, ``Voronoi-based
  multi-robot autonomous exploration in unknown environments via deep
  reinforcement learning,'' \emph{IEEE Transactions on Vehicular Technology},
  vol.~69, no.~12, pp. 14\,413--14\,423, 2020.

\bibitem{schoettler2020deep}
G.~Schoettler, A.~Nair, J.~Luo, S.~Bahl, J.~A. Ojea, E.~Solowjow, and
  S.~Levine, ``Deep reinforcement learning for industrial insertion tasks with
  visual inputs and natural rewards,'' in \emph{2020 IEEE/RSJ International
  Conference on Intelligent Robots and Systems (IROS)}.\hskip 1em plus 0.5em
  minus 0.4em\relax IEEE, 2020, pp. 5548--5555.

\bibitem{li2018off}
J.~Li, T.~Chai, F.~L. Lewis, Z.~Ding, and Y.~Jiang, ``Off-policy interleaved $
  q $-learning: Optimal control for affine nonlinear discrete-time systems,''
  \emph{IEEE transactions on neural networks and learning systems}, vol.~30,
  no.~5, pp. 1308--1320, 2018.

\bibitem{behzadan2019adversarial}
V.~Behzadan and A.~Munir, ``Adversarial reinforcement learning framework for
  benchmarking collision avoidance mechanisms in autonomous vehicles,''
  \emph{IEEE Intelligent Transportation Systems Magazine}, vol.~13, no.~2, pp.
  236--241, 2021.

\bibitem{alshiekh2018safe}
M.~Alshiekh, R.~Bloem, R.~Ehlers, B.~K{\"o}nighofer, S.~Niekum, and U.~Topcu,
  ``Safe reinforcement learning via shielding,'' in \emph{Proc. of the 32nd
  AAAI Conf. on Artificial Intelligence}, 2018.

\bibitem{jansen2018shielded}
N.~Jansen, B.~K{\"o}nighofer, S.~Junges, and R.~Bloem, ``Shielded
  decision-making in mdps,'' \emph{arXiv preprint arXiv:1807.06096}, 2018.

\bibitem{zhang2020robust}
H.~Zhang, H.~Chen, C.~Xiao, B.~Li, M.~Liu, D.~Boning, and C.-J. Hsieh, ``Robust
  deep reinforcement learning against adversarial perturbations on state
  observations,'' in \emph{Advances in Neural Information Processing Systems},
  vol.~33.\hskip 1em plus 0.5em minus 0.4em\relax Curran Associates, Inc.,
  2020, pp. 21\,024--21\,037.

\bibitem{bloomfield2020towards}
R.~Bloomfield, G.~Fletcher, H.~Khlaaf, P.~Ryan, S.~Kinoshita, Y.~Kinoshit,
  M.~Takeyama, Y.~Matsubara, P.~Popov, K.~Imai \emph{et~al.}, ``Towards
  identifying and closing gaps in assurance of autonomous road vehicles--a
  collection of technical notes part 1,'' \emph{arXiv preprint
  arXiv:2003.00789}, 2020.

\bibitem{robu_train_2018}
V.~Robu, D.~Flynn, and D.~Lane, ``Train robots to self-certify as safe,''
  \emph{Nature}, vol. 553, no. 7688, pp. 281--281, 2018.

\bibitem{JMLR:v16:garcia15a}
J.~Garc{{\'i}}a, Fern, and o~Fern{{\'a}}ndez, ``A comprehensive survey on safe
  reinforcement learning,'' \emph{Journal of Machine Learning Research},
  vol.~16, no.~42, pp. 1437--1480, 2015.

\bibitem{calinescu_efficient_2021}
R.~Calinescu, C.~Paterson, and K.~Johnson, ``Efficient {Parametric} {Model}
  {Checking} {Using} {Domain} {Knowledge},'' \emph{IEEE Transactions on
  Software Engineering}, vol.~47, no.~6, pp. 1114--1133, 2021.

\bibitem{zhao_probabilistic_2019}
X.~Zhao, V.~Robu, D.~Flynn, F.~Dinmohammadi, M.~Fisher, and M.~Webster,
  ``Probabilistic model checking of robots deployed in extreme environments,''
  in \emph{Proc. of the {AAAI} {Conf.} on {Artificial} {Intelligence}},
  vol.~33, Honolulu, Hawaii, USA, 2019, pp. 8076--8084.

\bibitem{berkenkamp2017safe}
F.~Berkenkamp, M.~Turchetta, A.~P. Schoellig, and A.~Krause, ``Safe model-based
  reinforcement learning with stability guarantees,'' \emph{arXiv preprint
  arXiv:1705.08551}, 2017.

\bibitem{huh2020safe}
S.~Huh and I.~Yang, ``Safe reinforcement learning for probabilistic
  reachability and safety specifications: A lyapunov-based approach,''
  \emph{arXiv preprint arXiv:2002.10126}, 2020.

\bibitem{mandlekar2017adversarially}
A.~Mandlekar, Y.~Zhu, A.~Garg, L.~Fei-Fei, and S.~Savarese, ``Adversarially
  robust policy learning: Active construction of physically-plausible
  perturbations,'' in \emph{2017 IEEE/RSJ Int. Conf. on Intelligent Robots and
  Systems (IROS)}.\hskip 1em plus 0.5em minus 0.4em\relax IEEE, 2017, pp.
  3932--3939.

\bibitem{bansal2017emergent}
T.~Bansal, J.~Pachocki, S.~Sidor, I.~Sutskever, and I.~Mordatch, ``Emergent
  complexity via multi-agent competition,'' in \emph{ICLR'18}, 2018.

\bibitem{kurach2020google}
K.~Kurach, A.~Raichuk, P.~Sta{\'n}czyk, M.~Zaj\c{a}c, O.~Bachem, L.~Espeholt,
  C.~Riquelme, D.~Vincent, M.~Michalski, O.~Bousquet \emph{et~al.}, ``Google
  research football: A novel reinforcement learning environment,'' in
  \emph{Proceedings of the AAAI Conf. on Artificial Intelligence}, vol.~34,
  no.~04, 2020, pp. 4501--4510.

\bibitem{katz2017reluplex}
G.~Katz, C.~Barrett, D.~L. Dill, K.~Julian, and M.~J. Kochenderfer, ``Reluplex:
  An efficient {SMT} solver for verifying deep neural networks,'' in
  \emph{CAV}, 2017, pp. 97--117.

\bibitem{RHK2018}
W.~Ruan, X.~Huang, and M.~Kwiatkowska, ``Reachability analysis of deep neural
  networks with provable guarantees,'' in \emph{IJCAI}, 2018, pp. 2651--2659.

\bibitem{fulton2018verifiably}
N.~Fulton, ``Verifiably safe autonomy for cyber-physical systems,'' Ph.D.
  dissertation, Ph.D thesis, Computer Science Department, Carnegie Mellon
  University, 2018.

\bibitem{bastani2018verifiable}
O.~Bastani, Y.~Pu, and A.~Solar-Lezama, ``Verifiable reinforcement learning via
  policy extraction,'' in \emph{Proc. of the 32nd Int. Conf. on Neural
  Information Processing Systems}, ser. NIPS'18.\hskip 1em plus 0.5em minus
  0.4em\relax Red Hook, NY, USA: Curran Associates Inc., 2018, p. 2499–2509.

\bibitem{zhu2019inductive}
H.~Zhu, Z.~Xiong, S.~Magill, and S.~Jagannathan, ``An inductive synthesis
  framework for verifiable reinforcement learning,'' in \emph{Proceedings of
  the 40th ACM SIGPLAN Conf. on Programming Language Design and
  Implementation}, 2019, pp. 686--701.

\bibitem{lillicrap2015continuous}
T.~P. Lillicrap, J.~J. Hunt, A.~Pritzel, N.~Heess, T.~Erez, Y.~Tassa,
  D.~Silver, and D.~Wierstra, ``Continuous control with deep reinforcement
  learning,'' in \emph{ICLR'16}, 2016.

\bibitem{sutton2018reinforcement}
R.~S. Sutton and A.~G. Barto, \emph{Reinforcement learning: An
  introduction}.\hskip 1em plus 0.5em minus 0.4em\relax MIT press, 2018.

\bibitem{kwiatkowska_probabilistic_2018}
M.~Kwiatkowska, G.~Norman, and D.~Parker, ``Probabilistic {Model} {Checking}:
  {Advances} and {Applications},'' in \emph{Formal {System} {Verification}:
  {State}-of the-{Art} and {Future} {Trends}}, R.~Drechsler, Ed.\hskip 1em plus
  0.5em minus 0.4em\relax Cham: Springer, 2018, pp. 73--121.

\bibitem{zhao_towards_2019}
X.~Zhao, M.~Osborne, J.~Lantair, V.~Robu, D.~Flynn, X.~Huang, M.~Fisher,
  F.~Papacchini, and A.~Ferrando, ``Towards integrating formal verification of
  autonomous robots with battery prognostics and health management,'' in
  \emph{Software {Engineering} and {Formal} {Methods}}, ser. {LNCS}, P.~C.
  Ölveczky and G.~Salaün, Eds., vol. 11724.\hskip 1em plus 0.5em minus
  0.4em\relax Cham: Springer, 2019, pp. 105--124.

\bibitem{gerasimou_undersea_2017}
S.~Gerasimou, R.~Calinescu, S.~Shevtsov, and D.~Weyns, ``{UNDERSEA}: an
  exemplar for engineering self-adaptive unmanned underwater vehicles,'' in
  \emph{{IEEE}/{ACM} 12th {Int.} {Symp.} on {Software} {Engineering} for
  {Adaptive} and {Self}-{Managing} {Systems}}, Buenos Aires, Argentina, May
  2017, pp. 83--89.

\bibitem{kwiatkowska_prism_2011}
M.~Kwiatkowska, G.~Norman, and D.~Parker, ``{PRISM} 4.0: {Verification} of
  probabilistic real-time systems,'' in \emph{Computer {Aided} {Verification}},
  ser. {LNCS}, G.~Gopalakrishnan and S.~Qadeer, Eds., vol. 6806.\hskip 1em plus
  0.5em minus 0.4em\relax Berlin, Heidelberg: Springer Berlin Heidelberg, 2011,
  pp. 585--591.

\bibitem{dehnert_storm_2017}
C.~Dehnert, S.~Junges, J.-P. Katoen, and M.~Volk, ``A {Storm} is coming: {A}
  modern probabilistic model checker,'' in \emph{Computer {Aided}
  {Verification}}, ser. {LNCS}, R.~Majumdar and V.~Kunčak, Eds., vol.
  10427.\hskip 1em plus 0.5em minus 0.4em\relax Cham: Springer, 2017, pp.
  592--600.

\bibitem{christiano2016transfer}
P.~Christiano, Z.~Shah, I.~Mordatch, J.~Schneider, T.~Blackwell, J.~Tobin,
  P.~Abbeel, and W.~Zaremba, ``Transfer from simulation to real world through
  learning deep inverse dynamics model,'' \emph{arXiv preprint
  arXiv:1610.03518}, 2016.

\bibitem{yu2018towards}
Y.~Yu, ``Towards sample efficient reinforcement learning.'' in \emph{IJCAI},
  2018, pp. 5739--5743.

\bibitem{name}
Robotis, ``Robotis(2019) turtlebot3 -- e-manual, waffle pi,'' [Online]
  https://emanual.robotis.com/docs/en/platform/turtlebot3/overview/. (Accessed
  on 02 August 2021).

\bibitem{yang2018active}
Y.~Yang, J.~Zhu, X.~Zhang, and X.~Wang, ``Active disturbance rejection control
  of a flying-wing tailsitter in hover flight,'' in \emph{2018 IEEE/RSJ
  International Conference on Intelligent Robots and Systems (IROS)}.\hskip 1em
  plus 0.5em minus 0.4em\relax IEEE, 2018, pp. 6390--6396.

\bibitem{seo2019robust}
H.~Seo, D.~Lee, C.~Y. Son, C.~J. Tomlin, and H.~J. Kim, ``Robust trajectory
  planning for a multirotor against disturbance based on hamilton-jacobi
  reachability analysis,'' in \emph{2019 IEEE/RSJ International Conference on
  Intelligent Robots and Systems (IROS)}.\hskip 1em plus 0.5em minus
  0.4em\relax IEEE, 2019, pp. 3150--3157.

\bibitem{epifani_model_2009}
I.~Epifani, C.~Ghezzi, R.~Mirandola, and G.~Tamburrelli, ``Model evolution by
  run-time parameter adaptation,'' in \emph{Proc. of the 31st {Int.} {Conf.} on
  {Software} {Engineering}}, ser. {ICSE} '09.\hskip 1em plus 0.5em minus
  0.4em\relax Washington, DC, USA: IEEE Computer Society, 2009, pp. 111--121.

\bibitem{calinescu_formal_2016}
R.~Calinescu, C.~Ghezzi, K.~Johnson, M.~Pezzé, Y.~Rafiq, and G.~Tamburrelli,
  ``Formal verification with confidence intervals to establish quality of
  service properties of software systems,'' \emph{IEEE Transactions on
  Reliability}, vol.~65, no.~1, pp. 107--125, 2016.

\bibitem{woods2017resilience}
D.~D. Woods, \emph{Resilience engineering: {Concepts} and precepts}.\hskip 1em
  plus 0.5em minus 0.4em\relax CRC Press, 2017.

\bibitem{quigley2009ros}
M.~Quigley, K.~Conley, B.~Gerkey, J.~Faust, T.~Foote, J.~Leibs, R.~Wheeler,
  A.~Y. Ng \emph{et~al.}, ``Ros: an open-source robot operating system,'' in
  \emph{ICRA workshop on open source software}, vol.~3, no. 3.2.\hskip 1em plus
  0.5em minus 0.4em\relax Kobe, Japan, 2009, p.~5.

\bibitem{koenig2004design}
N.~Koenig and A.~Howard, ``Design and use paradigms for gazebo, an open-source
  multi-robot simulator,'' in \emph{IEEE/RSJ Int. Conf. on Intelligent Robots
  and Systems}, vol.~3.\hskip 1em plus 0.5em minus 0.4em\relax IEEE, 2004, pp.
  2149--2154.

\end{thebibliography}
% \newpage
% \input{Response_Letter.tex}

\end{document}